\documentclass[acmsmall,authorversion,nonacm]{acmart}
\AtBeginDocument{%
  \providecommand\BibTeX{{%
    \normalfont B\kern-0.5em{\scshape i\kern-0.25em b}\kern-0.8em\TeX}}}

\usepackage[capitalise]{cleveref}
\usepackage{multirow}
\usepackage{natbib}
\usepackage{graphicx}
\usepackage{acro}
\usepackage{enumitem}
\usepackage{soul}
\usepackage[capitalise]{cleveref}
\usepackage{algorithm}
\definecolor{TableRowGray}{gray}{0.93}
\usepackage{balance}
\usepackage{colortbl}
\usepackage{tabularx}
\usepackage{setspace}
\usepackage{float}
\usepackage[font=footnotesize,labelfont=bf]{caption}
\newtheorem{problem}{Problem}
\newtheorem*{example}{Example}
\usepackage[para]{footmisc}
\usepackage{wrapfig}
\usepackage{algpseudocode}
\theoremstyle{plain}

\graphicspath{ {figures/} }
\def\BibTeX{{\rm B\kern-.05em{\sc i\kern-.025em b}\kern-.08em
    T\kern-.1667em\lower.7ex\hbox{E}\kern-.125emX}}
\usepackage[utf8]{inputenc}
\usepackage{pgfplots}
\usepackage{pgfplotstable}
\usepackage{filecontents}
\DeclareUnicodeCharacter{2212}{−}
\usepgfplotslibrary{groupplots,dateplot}
\usetikzlibrary{patterns,shapes.arrows}
\pgfplotsset{compat=newest}
\usepackage{graphicx}
\usepgfplotslibrary{colorbrewer}
\usetikzlibrary{pgfplots.statistics, pgfplots.colorbrewer,pgfplots.dateplot,
        shadows,
        matrix} 

\definecolor{blueLine}{RGB}{57,106,177}
\definecolor{blueFill}{RGB}{114,147,203}
\definecolor{redLine}{RGB}{204,37,41}
\definecolor{greenLine}{RGB}{0,250,0}
\definecolor{blackLine}{RGB}{0,0,0}
\definecolor{goldLine}{RGB}{160,82,45}
\definecolor{brightgreen}{rgb}{0.4, 1.0, 0.0}
\definecolor{brinkpink}{rgb}{0.98, 0.38, 0.5}
\definecolor{cadmiumyellow}{rgb}{1.0, 0.96, 0.0}
\definecolor{cinnamon}{rgb}{0.82, 0.41, 0.12}
\definecolor{darkorange}{rgb}{1.0, 0.55, 0.0}
\definecolor{darkspringgreen}{rgb}{0.09, 0.45, 0.27}
\usepackage[font=footnotesize]{caption}
\usepackage{xcolor}
\usepackage{multirow}

\newif\ifshowRevisions \showRevisionsfalse

\newcommand{\revision}[1]{%
\ifshowRevisions
{\leavevmode\color{blue}#1\ignorespaces%
}
\else
#1%
\fi
}

\usepackage{soul}
\usepackage[textsize=tiny, disable,  textwidth=1.23cm]{todonotes}

\newif\ifshowOld \showOldfalse

\newcommand{\oldad}[1]{%

}

\usepackage{listings}%
\definecolor{keywords}{HTML}{8A4A0B}

\definecolor{background}{HTML}{EEEEEE}

\definecolor{comments}{HTML}{868686}

\lstdefinelanguage{sdl}{
 morekeywords={scene,long,string,uniform,distribution,int,entity,'{','}'},
 keywordstyle=\color{keywords},
    basicstyle=\scriptsize\ttfamily,
 morecomment=[l]{//}, 
 morecomment=[s]{/*}{*/}, 
 morestring=[b]",
    basicstyle=\scriptsize\ttfamily,%
 commentstyle=\color{comments}\ttfamily,
numbers=right,
    numberstyle=\scriptsize,
    stepnumber=1,
    numbersep=8pt,
breaklines=true,
    frame=tb,
 tabsize=4}

\newcounter{reviewercomment}[section]

\definecolor{darkblue}{rgb}{0,0,0.5}

\DeclareAcronym{cps}{
  short = CPS,
  long = Cyber-Physical System,
}

\DeclareAcronym{cp}{
  short = CP,
  long = Change Point,
}

\DeclareAcronym{e2e}{
  short = e2e,
  long = end-to-end,
}

\DeclareAcronym{vae}{
  short = VAE,
  long = Variational Autoencoder,
}

\DeclareAcronym{bvae}{
  short = $\beta$-VAE,
  long = $\beta$-Variational Autoencoder,
}

\DeclareAcronym{icp}{
  short = ICP,
  long = Inductive Conformal Prediction,
}

\DeclareAcronym{bo}{
  short = BO,
  long = Bayesian Optimization,
}

\DeclareAcronym{kl}{
  short = KL-divergence,
  long = KL-divergence,
}

\DeclareAcronym{cusum}{
  short = CUSUM,
  long = Cumulative Sum,
}

\DeclareAcronym{gan}{
  short = GAN,
  long = Generative Adversarial Network,
}

\DeclareAcronym{lec}{
  short = LEC,
  long = Learning Enabled Component,
}

\DeclareAcronym{les}{
  short = LES,
  long = Learning Enabled Cyber-Physical System,
}

\DeclareAcronym{sdl}{
  short = SDL,
  long = Scenario Description Language
}

\DeclareAcronym{ood}{
  short = OOD,
  long = Out-of-Distribution
}

\DeclareAcronym{OOD}{
  short = OOD,
  long = Out-of-Distribution
}

\DeclareAcronym{mig}{
  short = MIG,
  long = Mutual Information Gap
}

\DeclareAcronym{ad}{
  short = AD,
  long = Anomaly Detection
}
\DeclareAcronym{elbo}{
  short = ELBO,
  long = Evidential Lower Bound
}

\DeclareAcronym{av}{
  short = AV,
  long = Autonomous Vehicle
}

\DeclareAcronym{dnn}{
  short = DNN,
  long = Deep Neural Network
}

\begin{document}

  
\title[Efficient Out-of-Distribution Detection Using Latent Space of $\beta$-VAE for Cyber-Physical Systems]{Efficient Out-of-Distribution Detection Using Latent Space of $\beta$-VAE for Cyber-Physical Systems}

\author{Shreyas Ramakrishna}
\email{shreyas.ramakrishna@vanderbilt.edu}
\affiliation{%
  \institution{Vanderbilt University}
}

\author{Zahra Rahiminasab}
\email{RAHI0004@e.ntu.edu.sg}
\affiliation{%
  \institution{Nanyang Technological University}
}

\author{Gabor Karsai}
\email{gabor.karsai@vanderbilt.edu}
\affiliation{%
  \institution{Vanderbilt University}
}

\author{Arvind Easwaran}
\email{arvinde@ntu.edu.sg}
\affiliation{%
  \institution{Nanyang Technological University}
}

\author{Abhishek Dubey}
\email{abhishek.dubey@vanderbilt.edu}
\affiliation{%
  \institution{Vanderbilt University}
}

\renewcommand{\shortauthors}{S. Ramakrishna et al.}

\begin{abstract}
\revision{Deep Neural Networks are actively being used in the design of autonomous Cyber-Physical Systems (CPSs). The advantage of these models is their ability to handle high-dimensional state-space and learn compact surrogate representations of the operational state spaces. However, the problem is that the sampled observations used for training the model may never cover the entire state space of the physical environment, and as a result, the system will likely operate in conditions that do not belong to the training distribution. These conditions that do not belong to training distribution are referred to as Out-of-Distribution (OOD). Detecting OOD conditions at runtime is critical for the safety of CPS. In addition, it is also desirable to identify the context or the feature(s) that are the source of OOD to select an appropriate control action to mitigate the consequences that may arise because of the OOD condition. In this paper, we study this problem as a multi-labeled time series OOD detection problem over images, where the OOD is defined both sequentially across short time windows (change points) as well as across the training data distribution. A common approach to solving this problem is the use of multi-chained one-class classifiers. However, this approach is expensive for CPSs that have limited computational resources and require short inference times. Our contribution is an approach to design and train a single $\beta$-Variational Autoencoder detector with a partially disentangled latent space sensitive to variations in image features. We use the feature sensitive latent variables in the latent space to detect OOD images and identify the most likely feature(s) responsible for the OOD. We demonstrate our approach using an Autonomous Vehicle in the CARLA simulator and a real-world automotive dataset called nuImages.}
\end{abstract}

\keywords{Cyber-Physical Systems, Deep Neural Networks, Out-of-Distribution, Disentanglement, $\beta$-Variational Autoencoders, Mutual Information Gap.}

\maketitle

\section{Introduction}
\label{sec:intro}
\revision{Significant advances in Artificial Intelligence (AI) and Machine Learning (ML) are enabling dramatic, unprecedented capabilities in all spheres of human life, including \acp{cps}. The fundamental advantage of AI methods is their ability to handle high-dimensional state-space and learn decision procedures or control algorithms from data rather than models. This is because high-dimensional real-world state spaces are complex and intractable for mathematical modeling and analysis. As such it is common to find AI components like \acp{dnn} in real-world \acp{cps} such as autonomous cars ~\cite{pomerleau1989alvinn,bojarski2016end}, autonomous underwater vehicles ~\cite{eski2014design}, and homecare robots ~\cite{ko2017neural}. However, there is still a gap in the safety and assurance of AI-driven \acp{cps} as shown by well known incidents in recent past ~\cite{vlasic2016self,kohli2019enabling}. 
The technical debt is fundamentally in the black-box nature of AI components, which hinders the use of classical software testing strategies (e.g., code coverage, function coverage). There is research progress on testing ~\cite{pei2017deepxplore,tian2018deeptest}, however, wide-spread applicability remains questionable. The problem is exacerbated due to the way the systems are being designed.

\textbf{The Safety Conundrum:} \ac{cps} design flows focus on designing a system $S$ that satisfies some requirements $R$ in an environment $E$. During the design process, the developer selects component models, each including a parameter vector and typed ports representing the component interface, from a repository and defines an architectural instance\footnote{An architectural instance of a design is a labeled graph where nodes are the ports of the components, and the edges represent interactions between the ports.} $A_S$ of the system such that $A_S \parallel E \models R$, while satisfying any compositional constraints across the component boundaries, often specified as pre-conditions and post-conditions ~\cite{dubey2011model}. The difficulty in this process is that in practice the environment is only approximated using a surrogate model $\hat{E}$ or a set of observations $|\Tilde{\hat{E}}|$ collected from real-world data. It is clear that $A_S \parallel \Tilde{\hat{E}} \models R$ does not imply $A_S \parallel E \models R$. In this sense, the system is being deployed with the assumption that $\Tilde{\hat{E}} \approx  E $. However, this is not a strong guarantee and could result in scenarios where the designed architecture may fail in the physical environment. Hence, runtime monitoring of the system is required to identify when $\Tilde{\hat{E}} \not\models E $ i.e., the observed samples are \ac{ood} with the real environment.} 

\begin{figure}[t!]
 \centering
 \includegraphics[width=0.8\columnwidth]{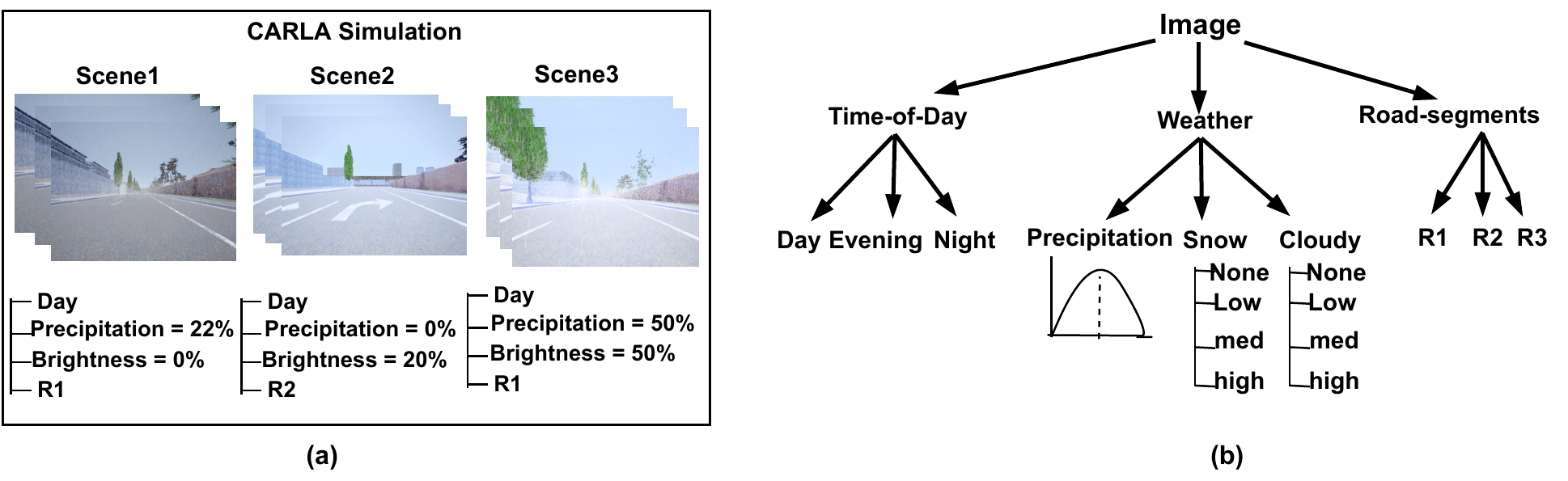}
\setlength{\abovecaptionskip}{2pt}
 \caption{\revision{\textbf{Scenes and feature representations}: (a) scenes with feature labels from CARLA simulation, (b) hierarchical representation of an image using its features. The features can take discrete or continuous values.}}
 \label{fig:image-features}
\vspace{-0.1in}
\end{figure}

\revision{To contextualize the problem, consider the case of a perception \ac{dnn} that consumes a stream of camera images to predict control actions (e.g., steer and speed) for autonomous driving tasks such as \acl{e2e} driving ~\cite{bojarski2016end}. In this context, the stream of images can be categorized as scenes. A scene is short time series of similar images contextualized by certain environmental features such as weather, brightness, road conditions, traffic density, among others, as shown in \cref{fig:image-features}. These features are referred to as semantic labels \cite{nuimage} or generative factors ~\cite{higgins2016beta}, and in this paper, we refer to them as features. These features can take continuous or discrete values, and these values can be sampled for generating different scenes as shown in \cref{fig:image-features}. The images from these scenes are collectively used for training the \ac{dnn}. These features effectively specify the context in which the system is operating and influence the sensitivity and correctness of the \ac{dnn}'s predictions, especially in cases where the features representing the scenes used for training do not cover all the values found in real-world. In this work, we primarily focus on perception \acp{dnn}, so we define the problem in terms of images. 

\textbf{Problem Definition:}
The problem, in this case, is to identify: (a) if the current image of the operational scene is \ac{ood} with respect to the training set, and (b) feature(s) likely responsible for the \ac{ood}. By this, we mean that if the training set used to train a \ac{dnn} did not include the scenes with heavy rain, then we want to identify during operation that the \ac{ood} is due to precipitation. In addition, as the images received by \acp{cps} are in time series, it is important to identify if the current image has changed with respect to the previous images in the time series. Identifying these changes is referred to as change point detection in literature. Change points in the values of a feature can increase the system's risk, as illustrated in our previous work ~\cite{hartsell2021resonate}. So, it is critical to identify these change points during operation. Formally, we summarize the problems as follows: \underline{\emph{Problem 1a}} - identify if the current image is \ac{ood} with respect to the training set. \underline{\emph{Problem 1b}} - identify the feature(s) most likely responsible for the \ac{ood}, and \underline{\emph{Problem 2}} - identify if the current image is \ac{ood} to the previous images in time series.} 

\textbf{State-of-the-art:} Multi-chained one-class classifiers \cite{tsoumakas2009mining} are commonly used for solving the multi-label anomaly detection problem. But the performance of these chains deteriorates in the presence of strong label correlation ~\cite{yeh2017learning}. Additionally, training one classifier for each label gets expensive for real-world datasets which have a large label set ~\cite{read2011classifier}. Another problem with traditional classifiers like Principal Component Analysis ~\cite{ringberg2007sensitivity}, Support Vector Machine ~\cite{scholkopf2001estimating}, and Support Vector Data Description (SVDD) ~\cite{wang2013modified} is that they fail in images due to computational scalability ~\cite{ruff2018deep}.

To improve the effectiveness of the classifiers, researchers have started investigating probabilistic classifiers like \ac{gan} ~\cite{goodfellow2014generative} and \ac{vae} ~\cite{kingma2013auto}. \ac{gan} has emerged as the leading paradigm in performing unsupervised and semi-supervised \ac{ood} detection ~\cite{akcay2018ganomaly,zenati2018efficient}, but problems of training instability and mode collapse ~\cite{creswell2018generative,vu2019anomaly} have resulted in \ac{vae} based methods being used instead. In particular, the \ac{vae} based reconstruction approach has become popular for detecting \ac{ood} data ~\cite{an2015variational,cai2020real}. However, this approach is less robust in detecting anomalous data that lie on the boundary of the training distribution ~\cite{denouden2018improving}. To address this, the latent space generated by a \ac{vae} is being explored ~\cite{denouden2018improving,vasilev1806q,sundar2020out}.
The latent space is a collection of latent variables ($L$), where each latent variable is a tuple defined by the parameters ($\mu$,$\sigma$) of a latent distribution ($z$) and a sample generated from the distribution. However, the traditional approach of just training a single \ac{vae} on all input data leads to unstructured and entangled distributions ~\cite{klys2018learning}, which makes the task of isolating the feature(s) responsible for \ac{ood} hard.

\revision{ \textbf{This paper}: Our approach in this work is to investigate use of latent space disentanglement for detecting \ac{ood} images used by perception \ac{dnn}s. This idea builds upon recent progress in structuring and disentangling the latent space ~\cite{bengio2013representation,higgins2016beta}. Effectively, the latent space generated by the encoder of a \ac{vae} is a Gaussian mixture model of several overlapping and entangled latent variables, each of which encodes information about the image features. However, as can be seen in \cref{fig:disentnagle}-a, the latent variables form a single large cluster which makes it hard to use them for \ac{ood} detection. Disentanglement is a state of the latent space where each latent variable is sensitive to changes in only one feature while being invariant to changes in the others ~\cite{bengio2013representation}. That is, the single large cluster of \cref{fig:disentnagle}-a is separated into several smaller clusters of single latent variables if the features are independent. Such disentangled latent variables have been successfully used in several tasks like face recognition ~\cite{tran2017disentangled,peng2017reconstruction}, video predictions ~\cite{hsieh2018learning}, and anomaly detection ~\cite{wang2020oiad}. However, disentangling all the latent variables is extremely hard for real-world datasets and is shown to be highly dependent on inductive biases ~\cite{locatello2018challenging} and feature correlations.  

\begin{figure}[t]
 \centering
 \includegraphics[width=\columnwidth]{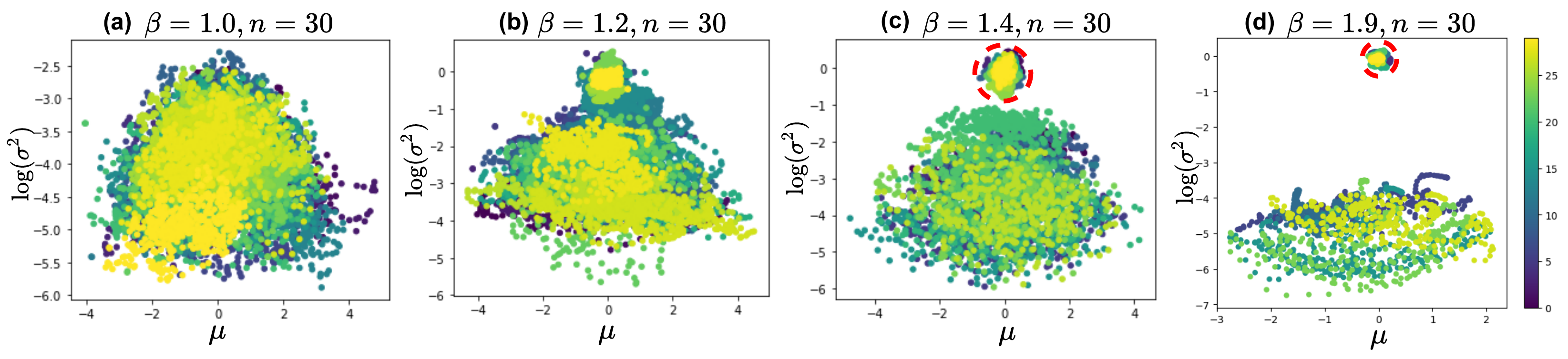}
\setlength{\abovecaptionskip}{-4pt}
 \caption{\textbf{Visualizing Latent Space Disentanglement}: Scatter plots illustrating the latent distributions ($\mu$, log($\sigma^2$)) generated using a $\beta$-VAE with different $\beta$ values. Each latent distribution in the latent space is represented using distinct color shades. CARLA images generated in \cref{sec:evaluation} was used to generate these latent distributions. For $\beta$=1, the generated latent distributions are entangled. For $\beta>1$, the latent distributions are partially disentangled with a few latent distributions (inside the red circle) encoding independent features moving close to $\mu$=0 and log($\sigma^2$)=0, and the others moving away. Plot axis: x-axis represents the mean of the latent distributions in the range [-5,5], and the y-axis represents the log of variance in the range [-5,5].}
 \label{fig:disentnagle}
\vspace{-0.11in}
\end{figure}

Nevertheless, even partially disentangling the latent variables can lead to substantial gains as shown by Jakab \emph{et al.}~\cite{jakab2020partial} and Mathieu \emph{et al.}~\cite{mathieu2019disentangling}. Partial disentanglement is a heuristic that groups the most informative latent variables into one cluster and the remaining latent variables that are less informative into another, as shown in \cref{fig:disentnagle}-c. This selective grouping enables better separation in the latent space for the train and test images as shown in \cref{fig:latent-plots} (see \cref{sec:evaluation}). However, the procedure for training a disentangled latent space for real-world \acp{cps} is hard. As a result, it is one of the aspects we focus on in this paper, along with interpreting the source of the \ac{ood}.} 

\textbf{Our Contributions}: \revision{We present an approach to generate a partially disentangled latent space and learn an approximate mapping between the latent variables and the image features to perform \ac{ood} detection and reasoning. The steps in our approach are data partitioning, latent space encoding, latent variable mapping, and runtime anomaly detection. To generate the partially disentangled latent space, we use a \ac{bvae}, which has a gating parameter $\beta$ that can be tuned to control information flow between the features and the latent space. For a specific combination of $\beta$ $>1$ and the number of latent variables ($n$), the latent space gets partially disentangled with few latent variables encoding most of the feature's information, while the others, encoding little information. We present a \acl{bo} heuristic to find the appropriate combination of the $\beta$ and $n$ hyperparameters. The heuristic establishes disentanglement as a problem of tuning the two hyperparameters. The selected hyperparameters are used to train a \ac{bvae} that is used by a latent variable mapping heuristic to select a set of most informative latent variables that are used for detection and identify the sensitivity of the latent variables towards specific features. The sensitivity information is used for reasoning the \ac{ood} images. Finally, the trained \ac{bvae} along with the selected latent variables are used at runtime for \ac{ood} detection and reasoning. We demonstrate our approach on two \ac{cps} case studies: (a) an \ac{av} in CARLA simulator ~\cite{dosovitskiy2017carla}, and (b) a real-world automotive dataset called nuImages \cite{nuimage}.}

\textbf{Outline}: The outline of this paper is as follows. We formulate the \ac{ood} detection problem in \cref{sec:ps}. \revision{ We introduce the background concepts in \cref{sec:background}}. We present our \ac{ood} detection approach in \cref{sec:approach}. We discuss our experiment setup and evaluation results in \cref{sec:evaluation}. \revision{Finally, we present related research in \cref{sec:RW} followed by conclusions in \cref{sec:conclusion}.}

\section{Problem Formulation}
\label{sec:ps}
\label{sec:problem}

\revision{To set up the problem, consider a \ac{cps} that uses a perception \ac{dnn} trained on image distribution $\mathcal{T}$, where $\mathcal{T}$ = \{$s_1,s_2, \dots,s_i$\} is a collection of scenes. A scene is a collection of sequential images \{$I_1,I_2, \dots,I_m$\} generated from a training distribution $P(\mathcal{T})$. Every image in a scene is associated with a set of discrete or continuous valued labels ($I\rightarrow2^\mathbb{L}$) belonging to the generative features of the environment (see \cref{fig:image-features}). It is important to note that the sampling rate of images depends on the dynamics of the system. With this model, we can define the problems we study as follows:}

\begin{problem}
\label{prob:ood-dectection}
Given a test image $I_t$, determine \textbf{(a)} if $I_t \in P(\mathcal{T})$, and \textbf{(b)} if ($I_t \not \in P(\mathcal{T})$) then identify the feature $f$ whose  $Label(I_t,f) \not \in P_{f}(\mathcal{T})$, where $P_{f}$ is the training distribution on the feature $f$.
\end{problem}

\begin{figure}[t]
\begin{tikzpicture}
\pgfmathsetlengthmacro\MajorTickLength{
      \pgfkeysvalueof{/pgfplots/major tick length} * 0.5
    }
\begin{groupplot}[group style={group size=3 by 1,horizontal sep = 0.8 cm, vertical sep = 1.2cm}]
\nextgroupplot[
  font=\footnotesize,
  width=0.3\columnwidth,
  height=0.25\columnwidth,
  xlabel=Time (s),
    legend style={at={(-0.27,-0.4)},
     anchor=north west},
ylabel style={align=center}, ylabel=Normalized \\ Steering Angle,
ymajorgrids,
xmajorgrids,
   xmin=0,
  xmax=20,
  ymin=-0.5,
  ymax=2.0,
    grid style={line width=.1pt, draw=gray!10},
    major grid style={line width=.2pt,draw=gray!50},
    ]
\addplot[solid, redLine] table[col sep=comma,x=t,y=s4]{results/csv-files/steer.csv};

\nextgroupplot[
  font=\footnotesize,
  width=0.3\columnwidth,
  height=0.25\columnwidth,
      xlabel=Time (s),
ymajorgrids,
xmajorgrids,
     xmin=0,
  xmax=20,
  ymin=-0.5,
  ymax=2.0,
 ]
 \addplot[solid,darkorange] table[col sep=comma,x=t,y=s3]{results/csv-files/steer.csv};

 \nextgroupplot[
  font=\footnotesize,
  width=0.3\columnwidth,
  height=0.25\columnwidth,
      xlabel=Time (s),
ymajorgrids,
xmajorgrids,
     xmin=0,
  xmax=20,
  ymin=-0.5,
  ymax=2.0,
 ]
 \addplot[dashed,blueLine] table[col sep=comma,x=t,y=s2]{results/csv-files/steer.csv};

 \end{groupplot}
\node[below right=1.3cm, font=\small\bf] (a) {(a)};
\node[right=2.2cm and 2.8 cm of a, font=\small\bf](b) {(b)};
\node[right=2.2cm and 2.9 cm of b, font=\small\bf] (T) {(c)};
\end{tikzpicture}
\setlength{\abovecaptionskip}{0pt}
\caption{\revision{\textbf{Problem illustration}: We trained an NVIDIA DAVE-II \ac{dnn} with images from scene1 and scene2 (\cref{fig:image-features}) to predict the steering values of an \ac{av} that travels on a straight road segment in CARLA simulator. We tested the network on three test scenes: (a) A scene that had precipitation and brightness values within the training distribution, (b) A scene with high precipitation (60\%) that was not in the training distribution. For this scene, the \ac{dnn} predictions deviate from the nominal value shown in the plot a, and (c) A scene where brightness value changed from low (20\%) to high (50\%) at $t=10$ seconds. The \ac{dnn} predictions were accurate until $t=10$ seconds, thereafter the predictions got erroneous.}}
\vspace{-0.8em}
\label{fig:weather}
\end{figure}
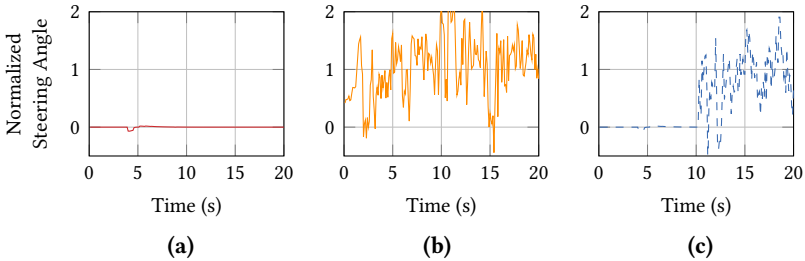

\begin{example}
To illustrate the problem, we trained an NVIDIA DAVE-II \ac{dnn} ~\cite{bojarski2016end} to perform \ac{e2e} driving of an \ac{av} in CARLA simulation. We trained the network on camera images from scene1 and scene2 (see \cref{fig:image-features}) to predict the steering control action for the \ac{av}. As shown in \cref{fig:weather}-a, the network's steering predictions were accurate when tested on images from training scenes. However, the predictions got erroneous when we used the network to predict on images of a new scene (scene3) with higher precipitation values outside the training distribution (\cref{fig:weather}-b). The error in steering predictions caused the \ac{av} crash of a sidewall. For this reason, if we knew that the precipitation level is compromising the network's predictions, we can switch to an alternative controller that operates on other sensor inputs (e.g., Radar or Lidar) rather than the camera images.
\end{example}

\begin{problem}
\label{prob:cp-detection}
\revision{Given the test image $I_t$ at time $t$, the goal is to determine if $I_t$ 
has changed with respect to the previous images in a time series window $(I[t-M+1],\dots,I[t])$, where $M$ is the window size.} 
\end{problem}

\begin{example}
\revision{To illustrate the problem, we use the same \ac{av} setup discussed in problem1. We tested the trained network on a new scene (scene4) with low brightness (in-distribution) for up to ten seconds, and the brightness was briefly high (\ac{ood}) for the next ten seconds. For this scene, the network predicted accurately for the first ten seconds and erroneously for the next ten seconds as shown in \cref{fig:weather}-c. Such an abrupt change in the image feature increases the \ac{av}'s risk of collision as demonstrated in our recent work \cite{hartsell2021resonate}. So, identifying the change points can be beneficial for reducing the system's risk of a consequence.}
\end{example}

\revision{\textbf{Detector Requirements}: We evaluate \ac{ood} detectors that solve these problems against the following properties.}

\vspace{-\topsep}
\begin{itemize}[noitemsep,leftmargin=*]
\item \revision{\underline{Robustness} - The detector should have low false positives and false negatives. A well known metric to measure robustness is F1-score = $(2 \cdot Precision \cdot Recall) \div (Precision + Recall)$. Precision is computed as $TP \div (TP + FN)$, and Recall is computed as $TP \div (TP + FP)$. Where TP is True positive, FP is False positive, and FN is false negative.}

\item \revision{\underline{Minimum Sensitivity (MS)} - The detector should have a minimum sensitivity  towards each feature ~\cite{martinez2008evolutionary}. For an image $I_t$ with feature labels $\mathbb{L}$=\{$l_1,l_2,\cdots l_k$\}, minimum sensitivity is defined as the minimum value of the detector's sensitivity for each feature label. $MS = min\{Si; i = 1,2,\cdots,k\}$, where $S_i$ is the sensitivity of the detector to the feature $i$. Recall has been a well known metric to measure the detector's sensitivity.}

\item \revision{\underline{Low Computation Overhead} - Our target platforms are resource constrained autonomous \acp{cps} like DeepNNCar \cite{ramakrishna2019augmenting}. Therefore, the detector should have a low resource signature.}

\item \revision{\underline{Low Execution time} - Autonomous \acp{cps} typically have a small sampling period (typically 50 to 100 milliseconds). Therefore, the detector should have a low execution time that is smaller than the system's sampling period.} 
\end{itemize}
\vspace{-\topsep}

\section{Background}
\label{sec:background}
In this section, we provide an overview of several basic concepts that are required to understand our \ac{ood} detection approach.

\subsection{Kullback-Leibler (KL) divergence}
\label{sec:kl}
\revision{\acl{kl} ~\cite{edwards2008elements} is a non-symmetric metric that can be used to measure the similarity between two distributions. For any probability distribution $p$ and $q$ the \ac{kl} can be computed as illustrated in \cref{eqn:kl-basic}. A \ac{kl} value close to zero indicates the two distributions are similar, while a larger value indicates their dissimilarity.

\begin{equation}
\small
D_{KL}(p||q)=\sum_{x \in \chi}p(x)log \frac{p(x)}{q(x)}
\label{eqn:kl-basic}
\end{equation}
\normalsize

Recently, the \ac{kl} metric is being utilized in several ways: (a) training loss function of different generative models (e.g., \ac{vae}) called the \ac{elbo} \cite{kingma2019introduction}, and (b) \ac{ood} detection metric \cite{vasilev1806q} that measures if the latent distributions generated by a generative model deviate from a standard normal distribution. The metric can be computed as shown below.

\begin{equation}
\small
KL(x) = D_{KL}(q_{\phi}(z_i|x)||\mathcal{N}(0,1))
\label{eqn:kl-metric}
\end{equation}
\normalsize

Where, $\mathcal{N}(0, 1)$ is a standard normal distribution. $q_{\phi}(z_i|x)$ is the distribution generated by the encoder of a \ac{vae} for a latent variable $L_i$, and input $x$.}


\subsection{\acl{bvae} ($\beta$-VAE)}
\acl{bvae} \cite{higgins2016beta} is a variant of the original \ac{vae} with a $\beta$ hyperparameter attached to the \ac{kl} (second term) of the \ac{elbo} loss function shown in \cref{eqn:elbo}. The network has an encoder that maps the input data ($x$) distribution $P(x)$ to a latent space ($z$) by learning a posterior distribution $q_{\phi}(z|x)$. The decoder then reconstructs a copy of the input data ($x'$) by sampling the learned distributions of the latent space. In doing so, the decoder also learns a likelihood distribution $p_{\theta}(x|z)$. \revision{The latent space is a collection of $n$ latent variables that needs to be selectively tuned in accordance with the input dataset. To remind, a latent variable is a tuple defined by the parameters ($\mu$,$\sigma$) of a latent distribution ($z$) and a sample generated from the distribution.}


\begin{equation}
ELBO(\theta,\phi,\beta;x,z) = {\mathbb{E}}_{q_{\phi}(z|x)}[log p_{\theta}(x|z)]-\beta D_{KL}(q_{\phi}(z|x)||p(z))
\label{eqn:elbo}
\end{equation}

$\theta$ and $\phi$ parameterize the latent variables of the encoder and the decoder, and $D_{KL}$ is the \acl{kl} metric discussed in \cref{sec:kl}. The first term computes the similarity between the input data $x$ and the reconstructed data $x'$. The second term computes the \ac{kl} between $q_{\phi}(z|x)$ and a predefined distribution $p_{\theta}(x|z)$, which is mostly the standard normal distribution $\mathcal{N}(0, 1)$.

\revision{\textbf{Tuning $\beta$ for disentanglement}: As suggested by Higgins \emph{et al.} ~\cite{higgins2016beta}, $\beta$ controls the amount of information that flows from the features to the latent space, and an appropriate hyperparameter combination of $\beta$ $>1$ and $n$ is shown to disentangle the latent space for independent features. \cref{fig:disentnagle} shows the latent space disentanglement for different values of $\beta$ while $n$ is fixed to $30$. For $\beta$ $=1$, the latent distributions are entangled. With $\beta$ $>1$, the latent space starts to partially disentangle with a few latent variables (inside the red circle) encoding most information about the features stay as a cluster close to $\mu$=0 and log($\sigma^2$)=0, and the others are uninformative and form a cluster that lies farther. However, as the $\beta$ value gets larger ($\beta$=1.9), the information flow gets so stringent that the latent space becomes uninformative ~\cite{mathieu2019disentangling}. So, finding an appropriate combination of $\beta$ and $n$ for disentanglement is a hard problem. To address this problem, we provide a heuristic in \cref{sec:encoding}.}

\subsection{\ac{mig}}
\label{sec:mig}
\acl{mig} is a metric proposed by Chen, Ricky TQ \emph{et al.} ~\cite{chen2018isolating} to measure the latent space disentanglement. It is an information theoretic metric that measures the mutual information between features and the latent variables. It measures the average difference between the empirical mutual information of the two most informative latent variables for each feature and normalizes this result by the entropy of the feature. \ac{mig} is computed using the following equation. 

\begin{equation}
\footnotesize
MIG=\frac{1}{|\mathcal{F}|}\sum_{f \in \mathcal{F}}\frac{1}{H(f)} (I_1(L_j;f)-I_2(L_j';f))
\label{eqn:mig}
\end{equation}

$I_1(L_j;f)$ represents the empirical mutual information between the most informative latent variable $L_j$ and the feature $f$. $I_2(L_j';f)$ represents the empirical mutual information between the second most informative latent variable $L_j'$ and the feature $f$. $H(f)$ is the entropy of information contained towards the feature $f$. In this work, we use \ac{mig} as a measure for selecting the right hyperparameter combination ($\beta$ and $n$) for the \ac{bvae}.

\begin{algorithm}[t]
    \algnewcommand\algorithmicforeach{\textbf{for each}}
    \algdef{S}[FOR]{ForEach}[1]{\algorithmicforeach\ #1\ \algorithmicdo}
    \footnotesize{}
	\caption{Computing MIG}
	\begin{flushleft}
	\textbf{Parameter}: number of latent variables $n$, image features $\mathcal{F}$, number of iterations $t$, trained \ac{bvae}\\
	\textbf{Input}: data partition $\mathcal{P}=\{P_1,P_2,....,P_m\}$\\
	\textbf{Output}: average MIG \\
	\end{flushleft}
	\begin{algorithmic}[1]
	\While{$i \leq t$}
	    \State Generate latent variable parameters ($\mu$, $\sigma$) using a trained \ac{bvae}
	    \ForEach {$P \in \mathcal{P} $}
	        \State Extract $\mu$, $\sigma$ parameters
	        \State Compute feature entropy H(f)
	        \State Compute latent variable entropies $H(L_j)$ and $H(L_j')$ 
	        \State Compute conditional entropies $H(L_j|f)$ and $H(L_j'|f)$
	        \State Compute Mutual Information of most informative latent variable $I_1(L_j;f) = H(L_j) - H(L_j|f)$
	        \State Compute Mutual Information of second most informative latent variable $I_2(L_j';f) = H(L_j') - H(L_j'|f)$
	   \EndFor
	   \State compute $MIG=\frac{1}{|\mathcal{P}|}\sum_{f \in \mathcal{P}}\frac{1}{H(f)} (I_1(L_j;f)-I_2(L_j';f))$
	   \State Append MIG to $L$
	\EndWhile
	\State \Return average MIG = sum($L$)/$t$
	\end{algorithmic} 
	\normalsize{}
\label{algo:MIG}
\vspace{-0.05in}
\end{algorithm}

\revision{\textbf{Implementation}: We have implemented \cref{algo:MIG} to compute \ac{mig}. Since \ac{mig} is computed based on entropy, the training set $\mathcal{T}$ should have images with feature labels that take different discrete and continuous values as shown in \cref{fig:image-features}. For} \revision{the ease of computation, we partition $\mathcal{T}$ into different partitions $\mathcal{P}$ using our approach discussed in \cref{sec:partition}. Our approach is to create partitions such that each partition will have images that have a variance in the value of a specific feature $f$, irrespective of the variance in the others. The feature with the highest variance represents the partition. Then, in each iteration, we use the feature representing the partition to compute the feature entropy, the conditional entropy, and the mutual information of the two most informative latent variables. The mutual information is then used to compute the \ac{mig} as shown in the algorithm. Finally, for robustness, we average the \ac{mig} across $t$ iterations.


\textbf{Complexity Analysis}: The complexity of the \ac{mig} algorithm in worst case is $O(t*|\mathcal{P}|*nq * |\mathcal{F}|*|\mathcal{T}|*n^2*ns)$. Where $t$ is the number of iterations, |$\mathcal{P}$| is the number of partitions, $n$ is the number of latent variables, |$\mathcal{F}$| is the number of features considered for the calculations, $nq$ is the number of unique values that each feature has in the partition (e.g., for a partition with brightness=10\%, and brightness=20\%, $nq$=2), $ns$ is the number of samples generated from each latent variable, and $|\mathcal{T}|$ is the size of training data. The specific values of these parameters for the \ac{av} example in CARLA simulation are $t=5$, $|p|=2$, $n=30$,  $\mathcal{F}=2$, $ns=500$, and $nq$=3.}

\subsection{\acl{icp} (ICP)}
\label{sec:icp}
\acl{icp} is a variant of the Conformal Prediction algorithm ~\cite{vovk2005algorithmic} that tests if a test observation ($x_t$) conforms to every observation in the training dataset ($\mathcal{T}$). However, comparing $x_t$ to every observation of $\mathcal{T}$ is expensive and gets complex with the size of $\mathcal{T}$. To address this, \ac{icp} splits $\mathcal{T}$ into two non-overlapping sets called as the proper training set ($\mathcal{T}_P$) which is used to train the prediction algorithm (e.g., \ac{dnn}) and the calibration set ($\mathcal{C}$) which is used to calibrate the test observations. In splitting the datasets, \ac{icp} performs a comparison of the $x_t$ to \revision{each element in the} $\mathcal{C}$ which is a smaller representative set of $\mathcal{T}$. The \ac{icp} algorithm has two steps: The first step involves, computing the non-conformity measure, which represents the dissimilarity between $x_t$ to the \revision{elements in the set} $\mathcal{C}$. The non-conformity measure is usually computed using conventional metrics like euclidean distance or K-nearest neighbors. But, in this work we use \ac{kl} as the non-conformity measure. The next step involves computing the p-value, which serves as evidence for the hypothesis that $x_t$ conforms to $\mathcal{C}$. \revision{Mathematically, the p-value is computed as the fraction of the observations in $\mathcal{C}$ that have non-conformity measure above the test observation $x_t$:  $p_{x_t} = |\{\forall \alpha \in \mathcal{C}|\alpha \geq \alpha_{x_t}\}|/|\mathcal{C}|$. Note for brevity, we drop the notation of $x$ and just use the term $p_t$ and $\alpha_t$.} Here $\alpha$ denotes the non-conformity measure for each observation in $\mathcal{C}$ and $\alpha_t$ denotes the non-conformity measure for $x_t$. 

Once $p_t$ is computed, it can be compared against a threshold $\tau$ $\in$ (0,1) to confirm if $x_t$ belongs to $\mathcal{T}$. However, such a threshold-based comparison is only valid if each test observation is i.i.d (independent and identically distributed) to $\mathcal{T}$, which is not true for \acp{cps} ~\cite{cai2020real}. Although the assumptions about i.i.d are not valid, \ac{icp} can still be applied under the weaker assumption of exchangeability. In our context, exchangeability means to test if the observations in $\mathcal{C}$ have the same joint probability distribution as the sequence of the test observations under consideration (are they permutations of each other). \revision{If they are, then we can expect the p-values to be independent and uniformly distributed in [0, 1] (Theorem 8.2, ~\cite{vovk2005algorithmic}), which can be tested using the martingale.} Exchangeability martingale~\cite{fedorova2012plug} has been used as a popular tool for testing the exchangeability and the i.i.d assumptions of the test observation with respect to $\mathcal{C}$. So, once the p-value for a test observation is computed, the simple mixture martingale~\cite{fedorova2012plug} can be computed as $\mathcal{M}_{t} = \int_{0}^{1} \left[ \prod_{i=1}^{t} \epsilon p_{i}^{\epsilon-1}\right] d\epsilon$.

\revision{Also, it is desirable to use a sequence of test observations rather than a single observation for improving the detection robustness. However, the observations received by \ac{cps} are in time series, which makes them non-exchangeable \cite{cai2020real}. The non-exchangeability nature hinders the direct application of the martingale to an infinitely long sequence of test observations. To address this, the authors in \cite{cai2020real} have suggested applying the martingale over a short window of the time series in which the test observations can be assumed to be exchangeable.  Then, the simple mixture martingale over a short time window $[t-M+1,t]$ of past $M$ p-values can be computed as $\mathcal{M}_{t} = \int_{0}^{1} \left[ \prod_{i=t-M+1}^{t} \epsilon p_{i}^{\epsilon-1}\right] d\epsilon$. The martingale will grow over time if and only if there are consistently low p-values within the time window, and the corresponding test observations are i.i.d. Otherwise, the martingale will not grow. It is important to note that the martingale computation is only valid for a short time window. The size of the window is dependent on the \ac{cps} dynamics, like the speed of the system. In our experiments, the system's speed was constant, so we used a fixed window size of $20$ images.} 


\subsection{\ac{cusum}} 
One of the problems we are dealing with is change detection. This problem is traditionally solved using \ac{cusum}  ~\cite{basseville1993detection}, which is a statistical quality control procedure used to identify variation based on historical data. It is computed as $S_0$ = 0 and $S_{t+1}$ = max(0, $S_t$+$x_t$-$\omega$), where $x_t$ is the sample from a process, $\omega$ is the weight assigned to prevent $S_t$ from consistently increasing to a large value. $S_t$ can be compared to a predefined threshold $\tau$ to perform the detection. $\omega$ and $\tau$ are hyperparameters that decide the detector's precision.
\begin{figure}[t]
 \centering
 \includegraphics[width=\columnwidth]{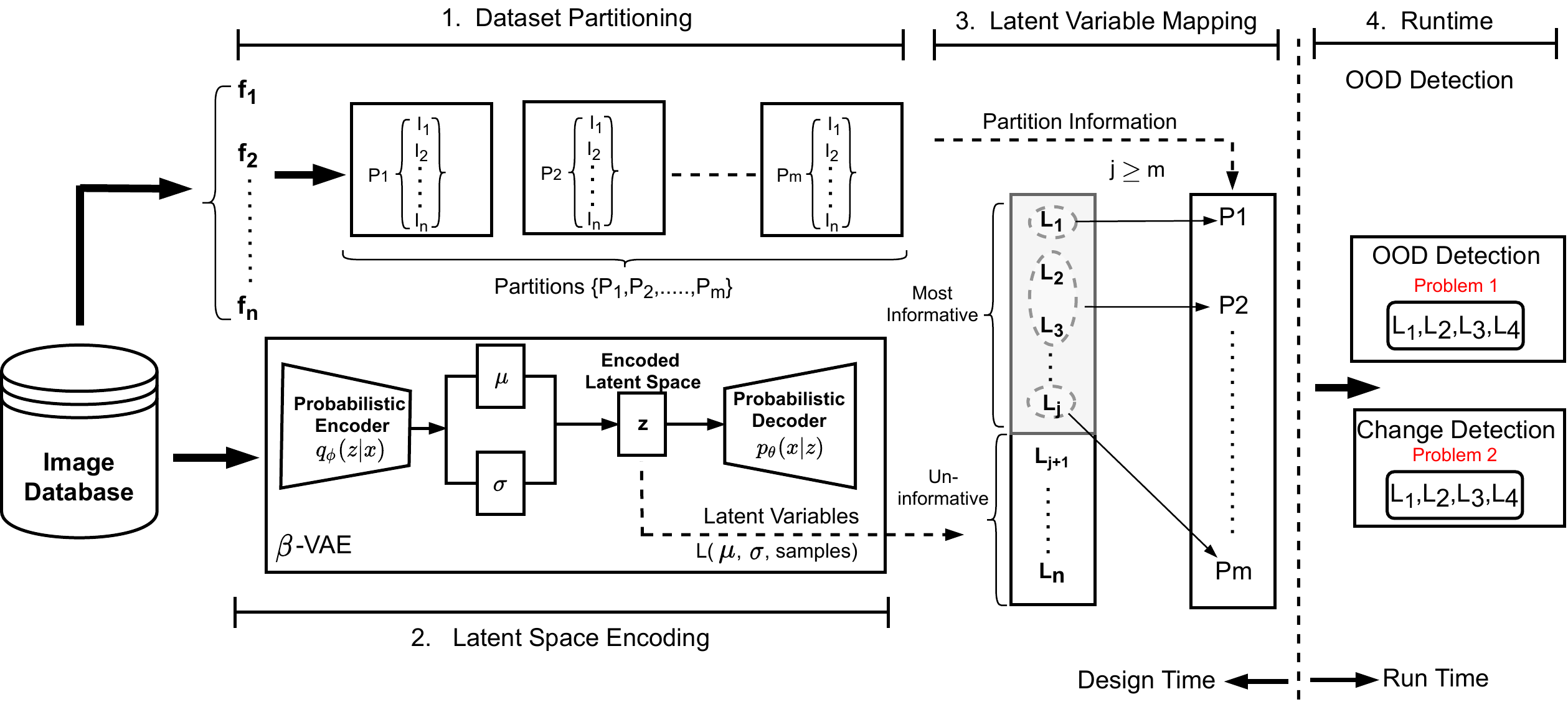}
\setlength{\abovecaptionskip}{0pt}
 \caption{\revision{\textbf{Detector design approach.} The steps of our approach include data partitioning, latent space encoding, latent variable mapping and runtime \ac{ood} detection.}}
 \label{fig:approach}
\vspace{-0.1in}
\end{figure}

\section{Our Approach}
\label{sec:approach-overview}
\label{sec:approach}

\revision{In this section, we present our approach that uses a \ac{bvae} to perform latent space based \ac{ood} detection and reasoning. Our approach is shown in \cref{fig:approach} and the steps involved work as follows: \underline{First}, we divide the multi-labeled datasets into partitions based on the variance in the feature values. A partition consists of images with one feature having higher variance in its values compared to others. \underline{Second}, we generate a partially disentangled latent space using a \ac{bvae}. As discussed earlier, the combination of $\beta$ and $n$ influence the level of latent space interpretability. To find the optimal combination, we propose a heuristic that uses the \acl{bo} algorithm ~\cite{snoek2012practical} along with \ac{mig} to measure disentanglement. \underline{Third}, we discuss a heuristic to perform latent variable mapping to identify the set of latent variables $\mathcal{L}_d$ that encodes most information of the image features. These latent variables are collectively used as the detector for \ac{ood} detection. Further, we perform a \ac{kl} based sensitivity analysis to identify the latent variable(s) $\mathcal{L}_f$ $\subseteq$ $\mathcal{L}_d$, that is sensitive to specific features. The latent variable(s) in $\mathcal{L}_f$ are used as reasoners to identify the \ac{ood} causing features. We discuss these steps in the rest of this section.}

\subsection{Data partitioning}
\label{sec:partition}

\revision{Data partitioning is one of the core steps of our approach. It is required for implementing the \ac{mig} and the latent variable mapping heuristics. We define a partition $P$ as a collection of images that have higher variance in the value of one feature as compared to the variances in the values of other features. To explain the concept of a partition, consider the features of images in training set $\mathcal{T}$ to be $\mathcal{F}=\{f_1,f_2,...,f_n\}$, and each feature can take a value either discrete or continuous as shown in \cref{fig:image-features}. We normalize these values for the ease of partitioning. Our goal is to group images in $\mathcal{T}$ into $m$ partitions $\mathcal{P}$ = $\{P_1,P_2,....,P_m\}$, such that a partition $P_j$ will have all the images with high variance in the values of feature $f_j$. For creating these partitions, we generate sub-clusters for each discrete valued feature(s) and sub-clusters for each continuous valued feature. In each of these sub-clusters, only the value of the feature under consideration $f_j$ changes while the value of other features remains unchanged. The clustering is done effectively through an agglomerative clustering algorithm. Thereafter, the partition for a feature is the union of all sub-clusters which can be represented as $P_j$ = $\{C_1 \cup C_2 .... \cup C_n\}$. It is important to note that each partition should have $variance>0$ in the value of the feature under consideration ($f_j$), irrespective of changes in the values of other features. To illustrate the partitioning concept, consider the example scenes in \cref{fig:image-features}. A precipitation partition for this example is a collection of images from two combinations, $C1$ = \{day, precipitation=22\%, brightness=0\%, R1\}, and $C2$ = \{day, precipitation=50\%, brightness=50\%, R1\}.

Our data partitioning approach works well for datasets with well defined labels that provide feature value(s). However, real-world automotive datasets such as nuScenes ~\cite{caesar2019nuscenes} and nuImages ~\cite{nuimage} provide semantic labels that are not always well defined, and they do not contain feature values. Also, they often have images in which several feature values change at once. Since the feature related information is not fully available, some prepossessing and thresholds selection for feature values are required for partitioning. Currently, the threshold selection for partitioning is performed by a human supervisor, but we want to automate it in the future. We have applied this partitioning technique on the nuScenes dataset in our previous work \cite{sundar2020out} and used it in this work for partitioning the nuImages dataset.}

\subsection{Latent Space Encoding}
\label{sec:encoding}
The second step of our design procedure is the selection and training of \ac{bvae} to generate a partially disentangled latent space encoding. However, the challenge is to determine the best combination of the $\beta$ and $n$ hyperparameters. To find this, we propose a novel greedy heuristic that formulates disentanglement as a hyperparameter search problem. The heuristic uses \acl{bo} (BO) algorithm with \ac{mig} as the objective function to maximize.

\textbf{Implementation}: The \ac{bo} algorithm builds a probability model of the objective function and uses it to identify the optimal model hyperparameter(s). \revision{The algorithm has two steps: a probabilistic Gaussian Process model that is fitted across all the hyperparameter points that are explored so far, and an acquisition function to determine which hyperparameter point to evaluate next} \cite{snoek2012practical}. We use these steps to search for an optimal hyperparameter for n $\in$ $N$, and $\beta$ $\in$ $B$. The heuristic using \ac{bo} algorithm is shown in \cref{algo:BO}, and the steps are discussed below.

\begin{algorithm}[!t]
    \footnotesize{}
    \caption{Bayesian Optimization Hyperparameter Selection}
	\begin{flushleft}
	\textbf{Parameter}: number of iterations $t$, initialization iterations $k$, explored list $\mathcal{X}$\\
	\textbf{Input}: training set $\mathcal{T}$, data partition $\mathcal{P}$, $N$, $B$\\
	\textbf{Output}: best $n$ and $\beta$\\
	\end{flushleft}
	\begin{algorithmic}[1]
	\For{$x=1,2,...,t$}
	    \If{$x \leq k$}
	        \State Randomly sample $n$, $\beta$ from $N$ and $B$
	    \Else
	         \State Find $n$, $\beta$ that optimizes the acquisition function over Gaussian Process
	    \EndIf
        \State Train \ac{bvae} on $\mathcal{T}$ using the selected $n$ and $\beta$
        \State Compute Average MIG 
        \State Append $n$, $\beta$ and MIG to $\mathcal{X}$ 
        \State Update the Gaussian Process posterior distribution using $\mathcal{X}$ 
	\EndFor
	\State \Return $n$ and $\beta$
	\end{algorithmic} 
	\normalsize{}
\label{algo:BO}
\vspace{-0.05in}
\end{algorithm}

\underline{First}, for $k$ initial iterations, we randomly pick values for $\beta$ and $n$ from the hyperparameter search space. The randomly selected n and $\beta$ combination is used to train a \ac{bvae} network and compute the \ac{mig} as discussed in \cref{algo:MIG}. The selected hyperparameters and the computed \ac{mig} are added to an explored list $\mathcal{X}$. After the initial iterations, the trained Gaussian Process model (the initial iterations are used to train and stabilizes the Gaussian process model) is fitted across all the hyperparameter points that were previously explored, and the marginalization property of the Gaussian distribution allows the calculation of a new posterior distribution g($x_n$) with posterior belief $\tilde{g}(x_n)$. Finally, the parameters ($\mu$, $\sigma$) of the resulting distribution are determined to be used by the acquisition function, which uses the posterior distribution to evaluate new candidate hyperparameter points.

\underline{Second}, an acquisition function is used to guide the search by selecting the hyperparameter(s) for next iteration. For this, it uses the $\mu$ and $\sigma$ computed by the Gaussian process. A commonly used function is the expected improvement (EI) ~\cite{jones1998efficient,gardner2014bayesian}, which can described as follows. Consider, $x_n$ to be some hyperparameter(s) point in the distribution g($x_n$) with posterior belief $\tilde{g}(x_n)$, and $x_{+}$ is the best hyperparameter(s) in $\mathcal{X}$ (explored list), then the improvement of the point $x_n$ is computed against $x_{+}$ as $I(x_n) = max\{0, (g(x_{+}) - g(x_n))\}$. Then, the expected improvement is computed as $EI(x_n) = \mathbb{E}\big[I(x_n)|x_n\big]$. Finally, the new hyperparameter(s) is computed as the point with the largest expected improvement as $x_{n+1} = argmax(EI(x_n))$. The new hyperparameter point is used to train a \ac{bvae} and compute the MIG, which is then added to the explored list $\mathcal{X}$. The list $\mathcal{X}$ is used to update the posterior distribution of the Gaussian process model in the next iteration. The two steps of the algorithm are iterated until maximum number of iterations is reached or can be terminated early if optimal hyperparameter(s) is consecutively selected by the algorithm for $j$ iterations (we chose $j$=3 in this work).

\revision{\textbf{Design Space Complexity}: Searching for optimal hyperparameters is a combinatorial problem that requires optimizing an objective function over a combination of hyperparameters. In our context, the \ac{bvae} hyperparameters to be selected are $\beta$ and $n$, and the objective function to optimize is the \ac{mig} whose complexity we have reported in the previous section. 
The range of the hyperparameters are n $\in$ $\{n_1,n_2,\dots,n_l\}$ and $\beta$ $\in$ $\{\beta_1,\beta_2,\dots,\beta_m\}$. The hyperparameter search space then becomes the Cartesian product of the two sets. In the case of the grid search, each point of this search space is explored, which requires training the \ac{bvae} and computing \ac{mig}. Grid search suffers from the curse of dimensionality since the number of evaluations exponentially grows with the size of the search space ~\cite{hutter2019automated}. In comparison, the random search and \ac{bo} algorithms do not search the entire space but search for a selected number of iterations $t$. While random search selects each point in the search space randomly, the \ac{bo} algorithm performs a guided search using a Gaussian process model and the acquisition function. 

In \ac{bo} algorithm, the first $k$ iterations stabilize the Gaussian Process model using randomly selected points in the space to train a \ac{bvae} and compute the \ac{mig}. After these iterations, the Gaussian process model is fitted to all the previously sampled hyperparameters. Then, the acquisition function based on expected improvement uses the posterior distribution of the Gaussian process model to find new hyperparameter(s) that may optimize the \ac{mig}. The new hyperparameter(s) are used to train a \ac{bvae} and compute the MIG. So, this process is repeated for $t$ iterations, and in every iteration, a \ac{bvae} is trained, and the \ac{mig} is computed as shown in \cref{algo:BO}. This intelligent search mechanism based on prior information makes the search technique efficient as it takes a smaller number of points to explore in the search space as compared to both grid search and random search ~\cite{snoek2012practical,hutter2019automated}. The \ac{bo} algorithm has a polynomial time complexity because the most time consuming operation is the Gaussian process which takes polynomial time ~\cite{hutter2019automated}. We report the experimental results for the three hyperparameter algorithms in \cref{Table:comptable}. As seen in the Table, the \ac{bo} algorithm takes the least time and iterations to select the hyperparameters that achieve the best \ac{mig} value as compared to the other algorithms.}

\begin{figure}[t!]
 \centering
 \includegraphics[width=0.8\columnwidth]{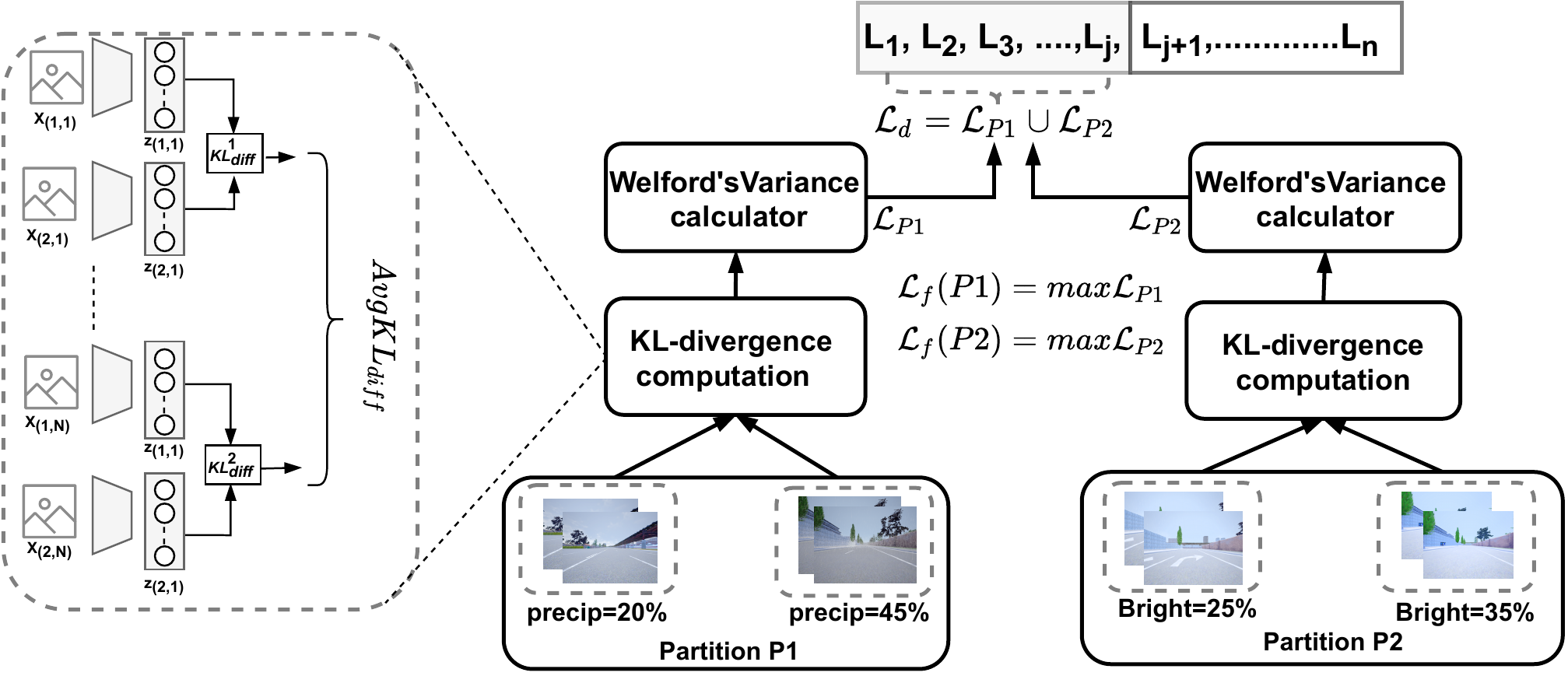}
 \caption{\revision{\textbf{Latent Variable Mapping}: Heuristic based on \ac{kl} and Welford's variance calculator to select latent variables for $\mathcal{L}_d$ and $\mathcal{L}_f$.}}
 \label{fig:latent}
\vspace{-0.1in}
\end{figure}

\subsection{Latent Variable Mapping}
\label{sec:approach_design_2}
Given the data partition set $\mathcal{P}$ and a trained \ac{bvae} that can generate the latent variable set $\mathcal{L}$, we find the most informative latent variables set $\mathcal{L}_d$ $\subseteq$ $\mathcal{L}$ that encodes information about the image features in the training set $\mathcal{T}$. As discussed earlier, the most informative latent variables form a separate cluster from the less informative ones when we partially disentangle the latent space using the \ac{bo} heuristic. Although the most informative latent variables are separately clustered in the latent space, we need a mechanism to identify them. To do this, we present a \ac{kl} based heuristic that is illustrated in \cref{fig:latent}.

\revision{The latent variables in $\mathcal{L}_d$ are used as detectors i.e., they can detect overall distribution shifts in the images (we discuss the exact procedure later). However, this does not solve the problem of identifying the specific feature(s) whose labels caused the \ac{ood}. For this, given the representative} \revision{feature $f$ of a partition has high variance, we identify the subset of latent variable(s) $\mathcal{L}_f \subseteq \mathcal{L}_d$ that is sensitive to the variations in $f$}. \revision{For example, if we have a partition with images that have different values in the precipitation (e.g., precipitation=20\%, precipitation=50\%, etc.), then we map latent variable(s) that are sensitive to changes in precipitation level.} The latent variable(s) in $\mathcal{L}_f$ is called the reasoner for feature $f$ and is used to identify if $f$ is responsible for the \ac{ood}. 
The steps in our heuristic are listed in \cref{algo:ld} and it works as follows: For each partition, $P \in \mathcal{P}$, and for each scene $s \in P$, we perform the following steps.

\begin{algorithm}[!t]
    \algnewcommand\algorithmicforeach{\textbf{for each}}
    \algdef{S}[FOR]{ForEach}[1]{\algorithmicforeach\ #1\ \algorithmicdo}
    \footnotesize{}
	\caption{Selecting Latent Variables for $\mathcal{L}_d$ and $\mathcal{L}_f$}
	\begin{flushleft}
	\textbf{Parameter}: global list $G$\\
	\textbf{Input}: data partitions $\mathcal{P}$ = $\{P_1,P_2,...,P_m\}$, number of latent variables $n$\\
	\textbf{Output}: $\mathcal{L}_d$ and $\mathcal{L}_f$ for each partition
	\end{flushleft}
	\begin{algorithmic}[1]
	\ForEach {$P \in \mathcal{P} $}
	    \ForEach {$s \in P $}
    	    \For{$l=1;\ l<= len(s);\ l=l+1$}
                \For{$i=0,1,2,.....,n$}
    	            \State $KL^i(x_l) = D_{KL}(q_{\phi}(z_i|x_l)||\mathcal{N}(0,1))$
    	            \State $KL^i(x_{l+1}) = D_{KL}(q_{\phi}(z_i|x_{l+1})||\mathcal{N}(0,1))$
    	           \State $KL^i_{l}(diff) = |KL^i(x_{l+1}) - KL^i(x_{l})| $
    	        \EndFor
    	  \EndFor
    	  \For{$i=1,2,....,n$}
            \State $Avg KL_{diff}^i = \frac{1}{len(s)-1} \sum_{l=1}^{len(s)-1}KL^i_{l}(diff) $
          \EndFor
        \State Store $Avg KL_{diff}^i$ to $G$ 
        \EndFor
	    \State Use Welford's algorithm to select $\mathcal{L}_P$, which is a set of $m$ latent variables with high variance in $Avg KL^i_{diff}$
	    \State $\mathcal{L}_f$ = max($\mathcal{L}_P$)
	\EndFor
	\State $\mathcal{L}_d$ = $\{\mathcal{L}_{P_1} \cup \mathcal{L}_{P_2} \cup.....\cup \mathcal{L}_{P_m}\}$
	\State \Return  $\mathcal{L}_d$, $\{\mathcal{L}_{f_1}, \mathcal{L}_{f_2},..., \mathcal{L}_{f_m}\}$ 
	\end{algorithmic}
	\footnotesize{}
\label{algo:ld}
\vspace{-0.05in}
\end{algorithm}

\underline{First}, we take two subsequent images $x_l$ and $x_{l+1}$ and pass each of them separately to the trained \ac{bvae} to generate the latent variable set  $\mathcal{L}$ for each of the images. \revision{As a remainder, $\mathcal{L}$ is a collection of $n$ latent variables, and each latent variable has a latent distribution ($z$) with parameters $\mu$ and $\sigma$.} Then, for each latent variable of the images ($x$) we compute a \ac{kl} between its latent distribution $q(z_i|x)$ and the standard normal distribution $\mathcal{N}(0,1)$ as discussed in \cref{sec:kl}. The computed \ac{kl} is $KL^i(x) = D_{KL}(q(z_i|x)||\mathcal{N}(0,1))$.   

\underline{Second}, we calculate the \ac{kl} difference between corresponding latent variables of the two images as: $KL^i_{l}(diff) = |KL^i(x_{l+1}) - KL^i(x_{l})| $. This procedure is repeated across all the subsequent images in the scene $s$. 

\underline{Third}, we compute an average \ac{kl} difference for each latent variable across all the subsequent images of the scene as follows.

\begin{equation}
\footnotesize
\revision{Avg KL_{diff}^i = \frac{1}{len(s)-1} \sum_{l=1}^{len(s)-1}KL^i_{l}(diff)}
\label{eqn:abs}
\end{equation}

where $KL^i_{l}(diff)$ is the \ac{kl} difference of the latent variable $L_i$ for the $l^{th}$ subsequent image pair in a scene $s$ and $len(s)$ is the number of images in $s$. This value indicates the average variance in the \ac{kl} value across each latent variable for all the images in the $s$. This approach of computing the variations across $s$ is motivated from the manual latent variable mapping technique in \cite{higgins2016beta}. Further, the $AvgKL^i_{diff}$ value is computed $\forall s \in P$. 

\revision{\underline{Fourth}, we then use the Welford's variance calculator ~\cite{welford1962note} to compute the variance in the $AvgKL^i_{diff}$ value across all the scenes in the partition. Welford's variance calculator computes and updates the variance in a single pass as the measurements are available. It does not require storing the measurements till the end for the variance calculation, which will make the variance calculation across several scenes faster. In our case, the variance calculator returns a partition latent variable set $\mathcal{L}_P$, which is a set of top $m$ latent variables that has the highest $AvgKL^i_{diff}$ across all images in $P$. Selecting an appropriate number of latent variables ($m$) for $\mathcal{L}_P$ is crucial, as we use it to select the latent variables for $\mathcal{L}_d$ and $\mathcal{L}_f$. The value for $m$ is chosen empirically, and the selection depends on the variances across the scenes of a partition. However, it is important to note that selecting a small value for $m$ may not include all the informative latent variables required for detection and choosing a large value for $m$ may include uninformative latent variables that may reduce the detection accuracy and sensitivity.   

Further, we choose the top latent variable(s) from $\mathcal{L}_P$ and use it as the reasoner ($\mathcal{L}_f$) for the partition. If the partition has variance in an independent image feature (e.g., brightness), then a single \revision{best latent variable} which has the most sensitivity can be used as the reasoner. However, if the feature is not independent, then more than one latent variable needs to be used, and the size of $\mathcal{L}_P$ increases. Besides, if two features correlate, then a single latent variable may be sensitive to both features. If such a latent variable is used for reasoning at runtime, and if it shows variation, then we attribute both the features to be responsible for the \ac{ood}.

\underline{Finally}, these steps are repeated for all the partitions in $\mathcal{P}$, and the latent variables for $\mathcal{L}_d$ is formed by $\{\mathcal{L}_{P_1} \cup \mathcal{L}_{P_2} \cup \dots \cup \mathcal{L}_{P_m}\}$. If the latent space is partially disentangled, then the top $m$ latent variables in each $\mathcal{L}_P$ will mostly be the same. Otherwise, the number of similar latent variables in each $\mathcal{L}_P$ will be small. }


\subsection{Runtime \acl{ood} Detection}
\label{sec:anomaly-detection}

\begin{figure}[t!]
 \centering
 \includegraphics[width=0.9\columnwidth]{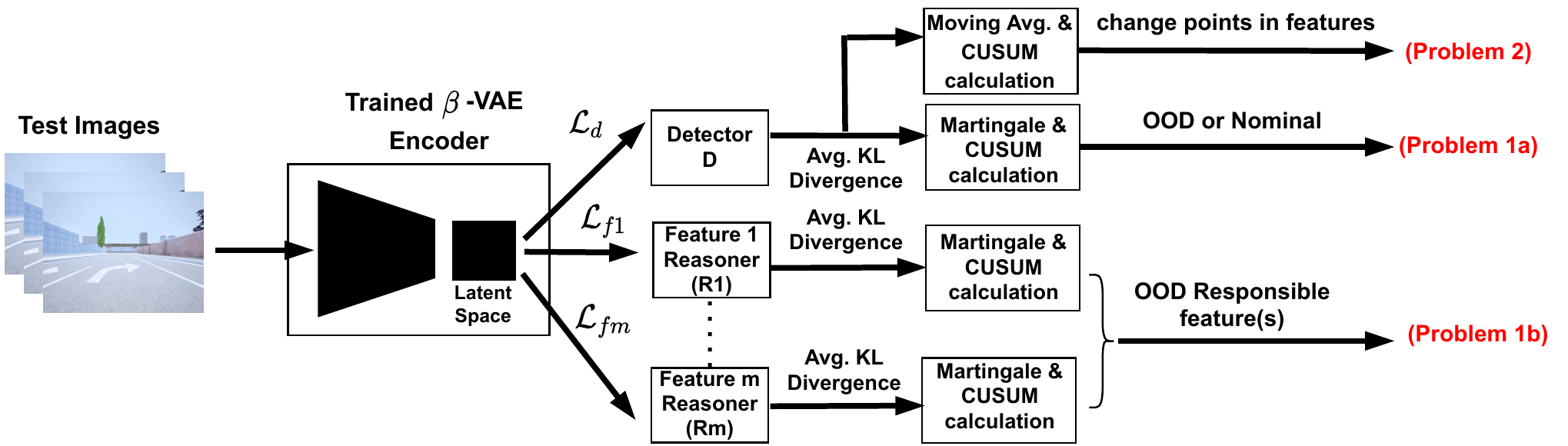}
 \caption{\revision{\textbf{Runtime \ac{ood} detection pipeline}: The trained \ac{bvae} detector at runtime provides three outputs that are sent to the decision manager to select an appropriate controller or control action that can mitigate the \ac{ood} problems as discussed in \cref{sec:problem}.}}
 \label{fig:pipeline}
\vspace{-0.1in}
\end{figure}


At runtime, we use the trained \ac{bvae} and the latent variable set $\mathcal{L}_d$ and $\mathcal{L}_f$ to detect \ac{ood} problems discussed in \cref{sec:problem}. 
\cref{fig:pipeline} shows the pipeline for runtime \ac{ood} detection, and it works as follows. As a test image $x_{t}$ is observed, the encoder of the trained \ac{bvae} is used to generate the latent space encoding. Then, the respective latent variables in ($\mathcal{L}_d$, $\mathcal{L}_f$) are sent to different processes to compute the average KL-divergence between the latent variables in $\mathcal{L}_d$ or latent variable(s) in $\mathcal{L}_f$ for the identified features. The \ac{kl} is computed between each latent variables in $\mathcal{L}_d$ and $\mathcal{L}_f$, and the normal distribution $\mathcal{N}(0,1)$ as shown in \cref{eqn:NM}. 

\begin{equation}
\alpha_t = KL(x_{t},\mathcal{N}(0,1)) = \frac{1}{L} \sum_{l=1}^{L}|D_{KL}(q(z_l|x_{t})||\mathcal{N}(0,1))|
\label{eqn:NM}
\end{equation}

Where $L$ is the number of selected latent distributions; for the detector, it is the number of latent variables in $\mathcal{L}_{d}$, and for each feature, it is the number of latent variables in $\mathcal{L}_{f}$. 

\revision{To detect the first \ac{ood} problem (\emph{Problem 1}), the \ac{kl} is used as the non-conformity score to compute the \ac{icp} and martingale score as shown in \cref{algo:algorithm1}. However, as the martingale can grow large very rapidly, we use the log of martingale. Then, a \ac{cusum} over the log of martingale is computed to identify when the martingale goes consistently high. The \ac{cusum} value $S_t$ of the detector latent variables $\mathcal{L}_d$ is compared against a detector threshold ($\tau_d$) to detect if the image $x_t$ is \ac{ood} compared to the calibration set. Then, the \ac{cusum} value $S_t$ of the reasoner latent variables $\mathcal{L}_f$ is compared against a reasoner threshold ($\tau_r$) to identify if the known feature(s) is responsible for the \ac{ood} as discussed in \emph{Problem 1b} of \cref{sec:problem}. The thresholds are empirically tuned as a tradeoff between false positives and mean detection delay~\cite{basseville1993detection}.

\begin{algorithm}[t]
\footnotesize{}
	\caption{$\beta$-VAE based OOD detection using ICP}
	\begin{flushleft}
	\textbf{Parameter}: sliding window length $M$, non-conformity measures of calibration set $\mathcal{C}$.\\
	\textbf{Input}: image $x_t$ at time $t$,  set of detector latent variables $\mathcal{L}_d$.\\
	\textbf{Output}: martingale score $\log \mathcal{M}_t$ at time $t$
	\end{flushleft}
	\begin{spacing}{1.0}
	\begin{algorithmic}[1]
	\State $\alpha_t = \sum_{\forall l \in \mathcal{L}_d}|D_{KL}(q(z_l|x_{t})||\mathcal{N}(0,1))|$
	\State $p_{t}=  \frac{\left (|\{\forall \alpha \in \mathcal{C} | \alpha \geq \alpha_{t} \} | \right )} {\left (|\mathcal{C}| \right )}$
	\State  $M_{t} = \int_{0}^{1} \left[ \prod_{i=t-M+1}^{t} \epsilon p_{i}^{\epsilon-1}\right] d\epsilon$
	\State \Return $\log \mathcal{M}_t$
	\end{algorithmic} 
	\end{spacing}
\normalsize{}
\label{algo:algorithm1}
\vspace{-0.3em}
\end{algorithm}

To detect the second \ac{ood} problem (\emph{Problem 2}), the latent variables in $\mathcal{L}_d$ is used with sliding window moving average and \ac{cusum} for change point detection. For this, the average \ac{kl} of all the latent variables in $\mathcal{L}_d$ is computed using \cref{eqn:NM}, and a moving average of the average \ac{kl} ($A_{KL}$) is computed over a sliding window [$x_{t-M+1}$,\dots,$x_t$] of previous $M$ images in the time series. $A_{KL}$ is used to compute the \ac{cusum} value $S_t$, which is compared against a threshold $\tau_{cp}$ to detect changes.} 


Finally, the outputs of the detector (See \cref{fig:pipeline}) are sent to the decision manager, which can use these detection results to perform system risk estimation \cite{hartsell2021resonate} or high level controller selection using simplex strategies \cite{seto1998simplex}. In this work, we use a simple decision logic that uses the \ac{ood} detection result to perform a control arbitration from the \ac{dnn} controller to an autopilot controller. 

\section{Experiments and Results}
\label{sec:evaluation}
We evaluate our approach using an \ac{av} example in the CARLA simulator \cite{dosovitskiy2017carla} and show the preliminary results from the real-world nuImages dataset \cite{nuimage}. The experiments\footnote{source code to replicate these experiments can be found at \url{https://github.com/scope-lab-vu/Beta-VAE-OOD-Detector}} in this section were performed on a desktop with AMD Ryzen Threadripper 16-Core Processor, 4 NVIDIA Titan Xp GPU's and 128 GiB memory.

\begin{figure}[!ht]
 \centering
 \setlength{\abovecaptionskip}{-2pt}
 \includegraphics[width=0.5\columnwidth]{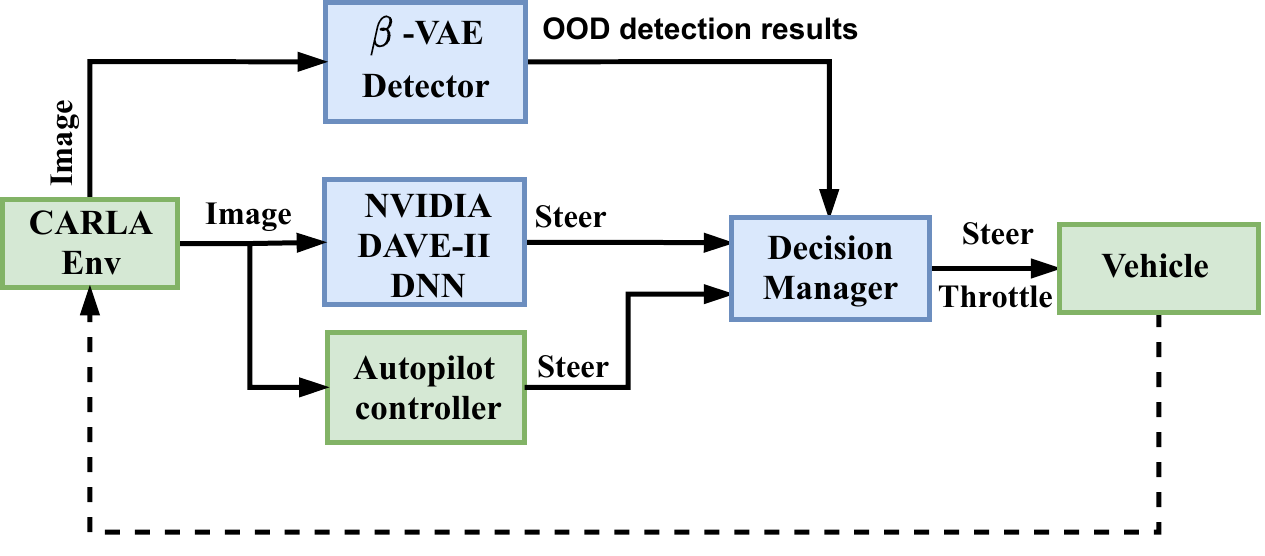}
 \caption{\textbf{\ac{av} block diagram}: The components in green come inbuilt with the CARLA simulator, while the components in blue are designed for our example. While the simulator requires a GPU, the other components are run on one CPU core to emulate a resource constrained setting.}
 \label{fig:carla-model}
\vspace{-0.1in}
\end{figure}

\subsection{System Overview}
\label{sec:system-setup}
Our first example system is an \ac{av} which must navigate different road segments in town1 of the CARLA simulator. The architecture of our \ac{av}, shown in \cref{fig:carla-model}, relies on a forward-looking camera for perception and a speedometer for measuring the system's speed. It uses the NVIDIA DAVE-II \ac{dnn} \cite{bojarski2016end} as the primary controller and the simulator's inbuilt autopilot mode as the secondary safety controller. In addition, a trained \ac{bvae} detector and reasoner are used in parallel to the two controllers. The detection results and the steering values from the two controllers are sent to a simplex decision manager, which selects the appropriate steering value for the system based on the detection result. That is, if the detector returns the image to be \ac{ood}, the decision manager selects the autopilot controller to drive the \ac{av}. The sampling period used in the simulation is 1/13 seconds, and the vehicle moves at a constant speed of $0.5$ m/s for all our experiments.   

\subsubsection{Operating modes} 
\begin{figure}[!h]
 \centering
 \includegraphics[width=\columnwidth]{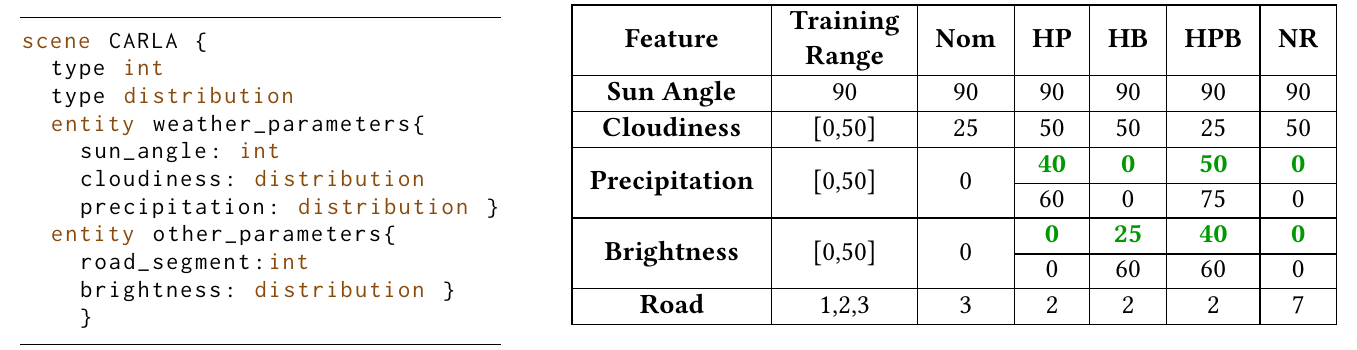}
 \caption{\revision{\textbf{Data Generation}: (left) A fragment of our \acl{sdl}. (right) Scenes in CARLA simulation: Nominal scene ($Nom$) is generated by randomly sampling the features in their training ranges. High Precipitation ($HP$), High Brightness ($HB$), High Precipitation \& Brightness ($HPB$) scene, and New Road ($NR$) are test scenes for which the feature values change from a training range value to a value outside the range at $t=20$ seconds. The initial values of precipitation and brightness are highlighted in green.}}
 \label{Table:Train}
\vspace{-0.1in}
\end{figure}

Our \ac{av} has two operating modes: (a) \textit{manual driving mode}, which uses CARLA's autopilot controller to drive around town1. The autopilot controller is not an AI component, but it uses hard-coded information from the simulator for safe navigation. We use this mode to collect the training set $\mathcal{T}$ and the test set ($\mathcal{T}_t$). These datasets are a collection of several CARLA scenes generated by a custom \ac{sdl} shown in \cref{Table:Train}; and (b) \textit{autonomous mode}, which uses a trained NVIDIA DAVE-II \ac{dnn} controller to drive the \ac{av}. In this setup, the \ac{bvae} detector is used in parallel to the \ac{dnn} controller to perform \ac{ood} detection.

\subsubsection{Data Generation}
\revision{Domain-specific \acp{sdl} such as Scenic ~\cite{fremont2019scenic} and MSDL ~\cite{msdl} are available for probabilistic scene generation. However, they did not fit our need to generate partition variations. Hence, we have implemented a simple \ac{sdl} in the textX ~\cite{dejanovic2017textx} meta language (See \cref{Table:Train}), which is combined with a random sampler over the range of the simulator's features like sun altitude, cloudiness, precipitation, brightness, and road segments to generate different scenes. 

\textbf{Train Scenes}: The training feature labels, and their values are shown in \cref{Table:Train}. These features were randomly sampled in the ranges shown in \cref{Table:Train} to generate eight scenes of $750$ images each that constituted the training set $\mathcal{T}$. Among these, two scenes had precipitation of 0\%, the brightness of $0\%$, sun angle $90$\textdegree, cloudiness of $25\%$, and road segment of $1$ and $2$ for each scene, respectively. Three scenes had different precipitation values (precip=$5\%$, precip=$40\%$, precip=$50\%$) while sun angle took a value of $90$\textdegree, cloudiness took a value of $25\%$, brightness took a value of $0\%$, and $10\%$, and the road segments took a value of $1$ and $2$. The remaining three scenes had different brightness values (bright=$9\%$, bright=$25\%$, bright=$40\%$), precipitation took a value of $0\%$ and $5\%$, and all the other parameters remained the same as the other scenes. We split $6000$ images of $\mathcal{T}$ into $4000$ images of $\mathcal{T}_P$ and $2000$ images of $\mathcal{C}$ in the standard 2:1 ratio (page 222 of \cite{hastie2009elements}) for \ac{icp} calculations. }

\begin{figure}[!ht]
 \centering
 \setlength{\abovecaptionskip}{0pt}
 \includegraphics[width=1.0\columnwidth]{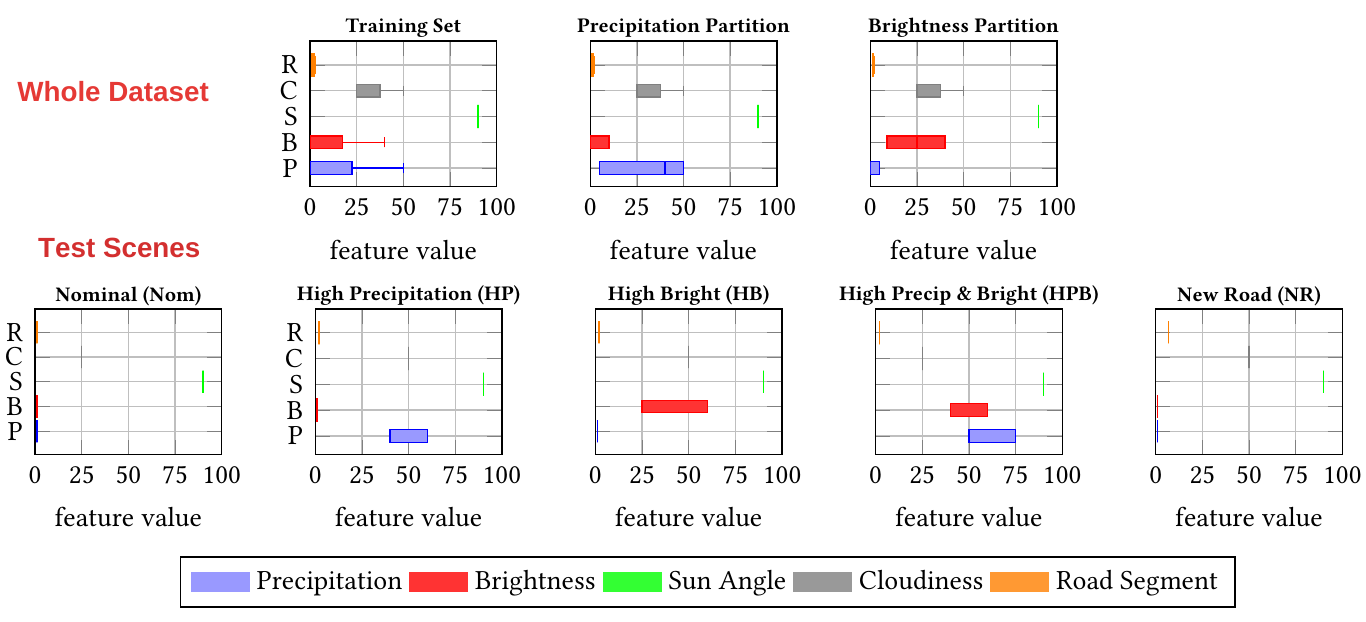}
 \caption{\revision{\textbf{Feature Value Variance:} Plots representing the normalized values of sun angle (S), cloudiness (C), precipitation (P), brightness (B), and road segments (R) for the training set, partitions, and the test scenes. We use the test scene abbreviations $Nom$ (Nominal), $HP$ (High Precipitation), $HB$ (High Brightness), $HPB$ (High Precipitation \& Brightness), and $NR$ (New Road segment) for the rest of the paper. The $NR$ scene has all the feature values like the $Nom$ scene, but it uses the road segment $7$ that is not in the training distribution.}}
 \label{fig:feature-variance}
\end{figure}

\textbf{Test Scenes}: The test scenes included a nominal scene ($Nom$) and four \ac{ood} scenes as shown in \cref{Table:Train}. Each scene was $20$ seconds long, and it had $260$ images. (1) Nominal Scene ($Nom$) was an in-distribution scene generated from the training distribution. (2) High Precipitation ($HP$) scene was an \ac{ood} scene in which the precipitation was increased from $40\%$ (in training range) to $60\%$ (out of training range) at $t=2$ seconds. (3) High Brightness ($HB$) scene was an \ac{ood} scene in which the brightness was increased from $25\%$ (in training range) to $60\%$ (out of training range) at $t=2$ seconds. (4) High Precipitation \& Brightness ($HPB$) scene was also an \ac{ood} scene in which both the precipitation and the brightness were increased out of the training range at $t=2$ seconds. (5) New Road ($NR$) scene was an \ac{ood} scene with a new road segment (segment=$7$) that was not in the training range, but the other features remained within the training range. \cref{fig:feature-variance} shows the variance in the normalized values of the scene features for the training set, partitions, and test scenes.

\subsection{$\beta$-VAE Detector}
\label{sec:detector-design}
We implement the steps of our approach discussed in \cref{sec:approach} to design a \ac{bvae} detector for our \ac{av} example.

\subsubsection{Data Partitioning}
\revision{Using our partitioning technique discussed in \cref{sec:partition}, we select three scenes with variance in precipitation values (precip=$5\%$, precip=$40\%$, precip=$50\%$) as the precipitation partition $P1$, and the remaining three scenes with variance in brightness values (bright=$9\%$, bright=$25\%$, bright=$40\%$) as the brightness partition $P2$.}

\subsubsection{Latent Space Encoding}
We applied the \ac{bo} algorithm based heuristic discussed in \cref{sec:encoding} to select and train the \ac{bvae} with appropriate hyperparameters.    


\textbf{Network Structure and Training}: We designed a \ac{bvae} network that has four convolutional layers $32/64/128/256$ with $5 x 5$ filters and $2 x 2$ max pooling followed by four fully connected layers with $2048$, $1000$, $250$  and $50$ neurons. A symmetric deconvolutional decoder structure is used as a decoder. This network along with images in $\mathcal{T}$ was used in the \ac{bo} algorithm discussed in \cref{algo:BO}. For each iteration of the \ac{bo} algorithm, the network was trained for $100$ epochs using the Adam gradient-descent optimizer and a two-learning scheduler, that had an initial learning rate $\eta$ = $1$ x $10$\textsuperscript{$-5$} for $75$ epochs, and subsequently fine-tuning $\eta$ = $1$ x $10$\textsuperscript{$-6$} for $25$ epochs. Learning rate scheduler is used to improve the model's accuracy and explore areas of lower loss. In addition, we had an early stopping mechanism to prevent the model from overfitting. 

\begin{table}[!t]
\footnotesize{}
\begin{tabular}{|c|c|c|c|c|c|}
\hline
\textbf{Algorithm} & \textbf{\# of Iterations} & \textbf{\begin{tabular}[c]{@{}c@{}}Iterations to \\ reach optimum\end{tabular}} & \textbf{\begin{tabular}[c]{@{}c@{}}Search Time\\ (min)\end{tabular}} & \textbf{max MIG} & \textbf{Selected Parameter} \\ \hline
{Grid}      & 360             & 5                                                                               & 10924.55                                                              & 0.0017           & 30,1.4                  \\ \hline
{Random}    & 50              & 40                                                                              & 1199.51                                                               & 0.00032          & 40,1.5                  \\ \hline
{\color[HTML]{009901}\textbf{BO}}   & \color[HTML]{009901}\textbf{50}              & \color[HTML]{009901}\textbf{16}                                                                              & \color[HTML]{009901}\textbf{837.05}                                                                & \color[HTML]{009901}\textbf{0.0018}           & \color[HTML]{009901}\textbf{30,1.4}                  \\ \hline
\end{tabular}
\caption{\textbf{Comparing Hyperparameter search algorithms}: \acl{bo} algorithm is compared against random and grid search algorithms.}
\label{Table:comptable}
\normalsize{}
\vspace{-0.2in}
\end{table}


In addition to network training, the algorithm also involved computing the \ac{mig} in every iteration. For computing it, we utilized the images and labels from partitions $P1$ and $P2$, which were generated in the previous step. To obtain robust \ac{mig}, we computed the latent variable entropy by randomly sampling $500$ samples from each latent variable in the latent space. To back this, we also averaged the \ac{mig} across five iterations. 


\textbf{Performance Comparison}: We compared the performance of \ac{bo}, grid, and random search algorithms. The results of these algorithms are shown in \cref{Table:comptable}. Random search and \ac{bo} algorithm was run for $50$ trials, while the grid search was run for $360$ trials across all combinations of n $\in$ $[30,200]$ and $\beta$ $\in$ $[1,5]$. In comparison, the \ac{bo} algorithm achieved the highest \ac{mig} value of $0.0018$ for $\beta=1.4$ and $n=30$ hyperparameters. It also took the shortest time of $837.05$ minutes (early termination because optimal hyperparameters(s) were found) as compared to the other algorithms.

\begin{table}[!th]
\footnotesize{}
\resizebox{0.5\columnwidth}{!}{%
\setlength\extrarowheight{2pt}
\begin{tabular}{|c|c|c|c|c|}
\hline
Partition & \multicolumn{4}{c|}{Latent Variable set $\mathcal{L}_P$}                           \\ \hline
P1        & \color[HTML]{009901}\textbf{$L_2$(0.09)}  & $L_{25}$(0.06) & $L_{0}$(0.05)  & $L_{29}$(0.02) \\ \hline
P2        & \color[HTML]{009901}\textbf{$L_{25}$(0.16)} & $L_{0}$(0.09)  & $L_{20}$(0.07) & $L_{2}$(0.07)  \\ \hline
\end{tabular}
}
\caption{\revision{\textbf{Latent Variable Mapping}: Ordered List of latent variable set $\mathcal{L}_P$ for $P1$ and $P2$. The latent variable $L_{2}$ had highest \ac{kl} variance across the scenes in $P1$. $L_{25}$ had the highest \ac{kl} variance across the scenes in $P2$. $\mathcal{L}_d$ is chosen as the union of the two $\mathcal{L}_P$ sets. Chosen $\mathcal{L}_d$ = $\{L_{0},L_{2},L_{20},L_{25},L_{29}\}$}}
\label{Table:avgkl}
\normalsize{}
\vspace{-0.2in}
\end{table}


\subsubsection{Latent Variable Mapping}
\label{sec:diagnosers}
We used the selected and trained \ac{bvae} along with the data partitions $P1$ and $P2$ to find latent variables for \ac{ood} detection ($\mathcal{L}_d$) and reasoning ($\mathcal{L}_f$). For each scene in the partitions, we applied the steps in \cref{algo:ld} as follows. \underline{First}, we used the successive images in each scene to generate a latent variable set $\mathcal{L}$ and then computed a \ac{kl} value. \underline{Second}, we computed an average \ac{kl} difference between corresponding latent variables of the two images. \underline{Third}, we computed the average \ac{kl} difference (using \cref{eqn:abs}) for each latent variable across all the subsequent images in a scene. We repeated these steps for all the scenes in both partitions. \underline{Finally}, for each partition we identified a partition latent variable set $\mathcal{L}_P$ using the Welford's algorithm as discussed in \cref{sec:approach_design_2}. The number of latent variables $m$ in $\mathcal{L}_P$ requires selection based on the dataset. In this work, the value for $m$ is chosen by human judgment. For the CARLA dataset, we chose the value of $m$ to be $4$. 

\begin{figure}[t]
 \centering
 \setlength{\abovecaptionskip}{0pt}
 \includegraphics[width=\columnwidth]{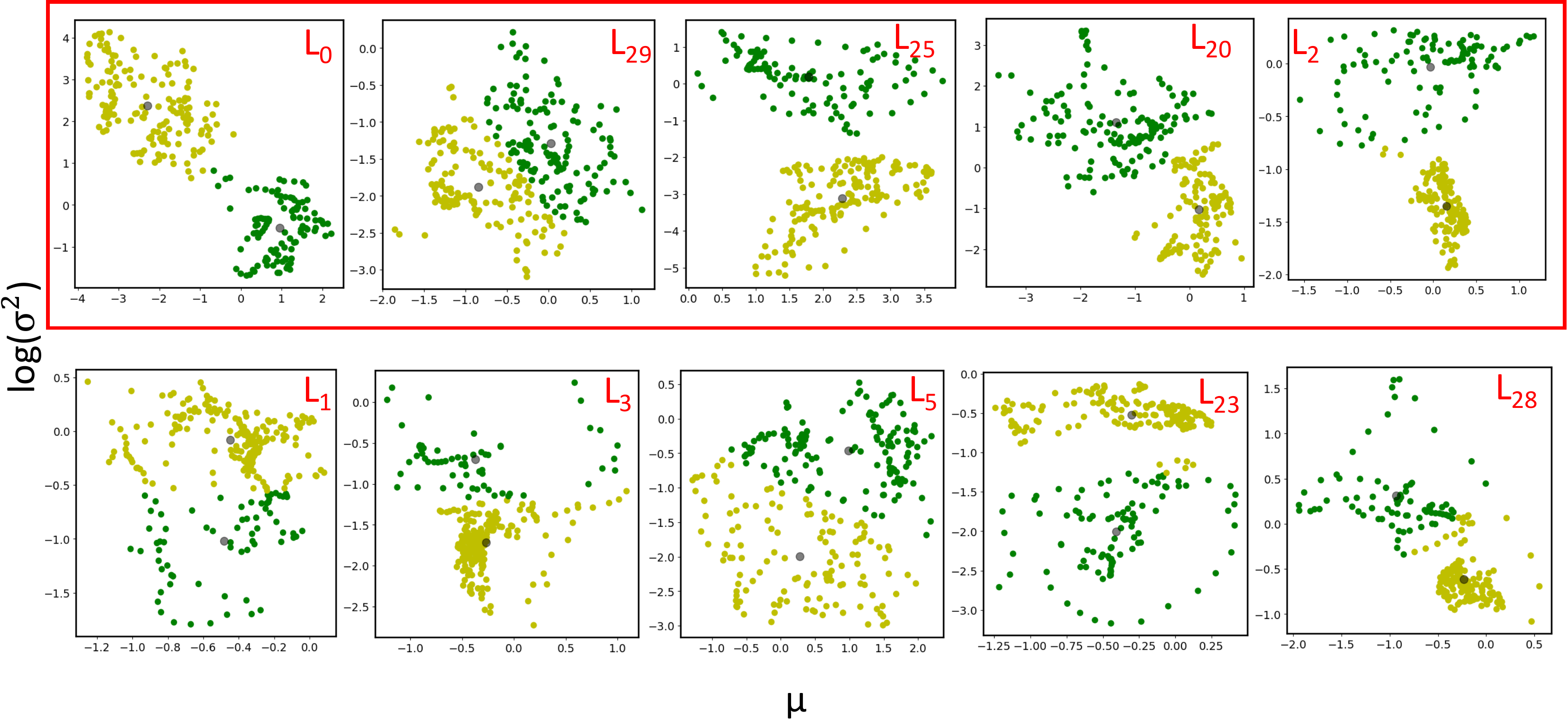}
 \caption{\revision{\textbf{Scatter plots of individual latent variables}: The latent distributions of the selected latent variables (highlighted in red) as compared to $5$ other latent variables that were not selected. Yellow points represent the distributions of the train images, and the green points represent the distributions of the test images that were \ac{ood}. The selected latent variables into a well-formed cluster, and there is a higher separation between the train and the test image clusters. The un-selected latent variables do not form clean clusters. Using the $5$ most informative latent variables for detection resulted in better robustness and minimum sensitivity as compared to using all the latent variables as shown in \cref{Table:design-space}. Plot axis: x-axis represents the mean of the latent distributions in the range $[-5,5]$, and the y-axis represents the log of variance in the range $[-5,5]$.}}
 \label{fig:latent-plots}
\vspace{-0.1in}
\end{figure}

\revision{Implementing these steps, we selected the partition latent variable sets $\mathcal{L}_{P1}$ = $\{L_2,L_{25},L_{20},L_0,L_{29}\}$ for $P1$ and $\mathcal{L}_{P2}$ = $\{L_{25},L_{0},L_{20},L_{2},L_{29}\}$ for $P2$. These latent variables had the highest \ac{kl} variance values and hence were selected. The \ac{kl} variance values of these latent variables are reported in \cref{Table:avgkl}. Then, the union of these two partition sets were used as the total detector $\mathcal{L}_d$ = $\{L_{0},L_{2},L_{20},L_{25},L_{29}\}$. \cref{fig:latent-plots} shows the scatter plots of the latent distributions of the selected latent variables and $5$ randomly selected latent variables ($L_1,L_3,L_5,L_{15},L_{28}$) for the train images (yellow points) and the test images (green points), which are \ac{ood} with high brightness. The latent distributions of the selected latent variables highlighted in the red box form evident clusters between the train images and the test images, and these clusters have a good intra-cluster separation. But, for the other latent variables, the distributions are scattered and do not form clean clusters, and these clusters are not well separated. 

Further, we chose one latent variable with the maximum variance in the partition latent variable sets and, it was used as the reasoner for that partition. We chose $L_2$ as the reasoner for the precipitation partition $P1$ and $L_{25}$ is used as the reasoner for the brightness $P2$. Our decision of choosing only one latent variable for reasoning is backed by the fact that the dataset was synthetically generated, and the features were not highly correlated. However, real-world datasets may require more than one latent variable for reasoning.}

\subsection{\acl{OOD} Detection Results from CARLA Simulation}
\label{sec:experiments}

\subsubsection{Evaluation Metrics} (1) \textit{Precision (P)} is a fraction of the detector identified anomalies that are real anomalies. It is defined in terms of true positives (TP), false positives (FP), and false negatives (FN) as $TP \div (TP + FN)$.  (2) \textit{Recall (R)} is a fraction of all real anomalies that were identified by the detector. It is calculated as $TP \div (TP + FP)$. The FP and FN for the test scenes is shown in \cref{Table:metric-table}. (3) \textit{F1-score} is a measure of the detector's accuracy that is computed as $2\times (P\times R) \div (P + R)$. (4) \textit{Execution Time} is computed as the time the detector receives an image to the time it computes the log of martingale. (5) \textit{Average latency} is the number of frames between the detection and the occurrence of the anomaly. (6) \textit{Memory Usage} is the memory utilized by the detector.

\begin{figure*}[!ht]
\setlength{\abovecaptionskip}{0pt}
    \centering
    \includegraphics[width=0.95\textwidth]{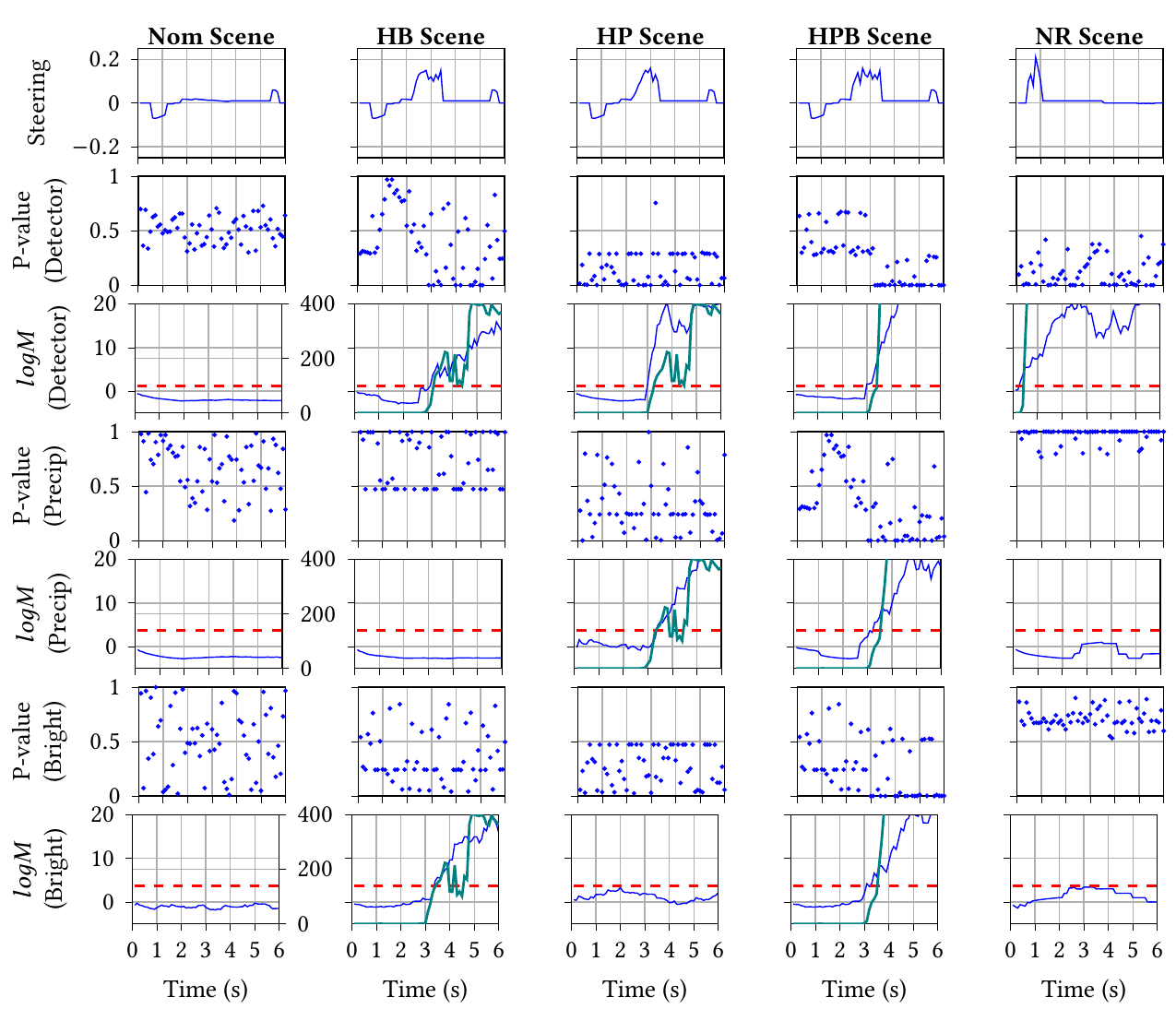}
    \caption{\textbf{Runtime \ac{ood} Detection}: Performance of the \ac{bvae} detector for the $5$ test scenes generated in \cref{sec:system-setup}. The solid blue line represents the log of martingale, the solid green lines represent the \ac{cusum} values and the dotted red lines represent the threshold ($\tau$) for \ac{cusum} comparison. Further, the left y-axis shows the log of martingale with range $[-5,20]$, and the right y-axis shows the \ac{cusum} value with range $[0,400]$. The cusum threshold represented in the red dotted lines are plotted to the right y-axis.}
    \vspace{-0.2in}
    \label{fig:confidence}
\end{figure*}

\subsubsection{Runtime OOD Detection}
\label{sec:runtime-ood}
\revision{We evaluate the performance of the selected \ac{bvae} network ($\beta=1.4$ and $n=30$) for the $5$ test scenes described in \cref{Table:Train}. Additional hyperparameters that were used by the detectors and the reasoners are as follows. The martingale sliding window size $M=20$, the \ac{cusum} parameters for the detector are $\omega_d=14$ and $\tau_d=100$, and for the reasoners are $\omega_r=18$ and $\tau_r=130$. These hyperparameters were selected empirically based on the false positive results from several trial runs. \cref{fig:confidence} summarizes the detector's performance for a short segment of the $5$ test scenes. For the $HP$, $HB$, $HPB$, and $NR$ scenes, the scene shifts from in-distribution to \ac{ood} at $t=2$ seconds, and they are used to illustrate the detection and reasoning capability of our approach. 

The $Nom$ scene has all the feature values within the training distribution, and it is used to illustrate the detector's ability to identify in-distribution images. As seen in \cref{fig:confidence}, the martingale of the detector and both the reasoners remain low throughout. In $HP$ and $HB$ scenes, the precipitation and the brightness feature values increase out of the training distribution at $t=2$ seconds. In the $HP$ scene, the martingales of the detector and the reasoner for the precipitation feature increase above the threshold after $t=2$ seconds. However, as seen martingale of the reasoner for brightness does not increase. So, we conclude that the precipitation feature is the reason for the \ac{ood}. In the $HB$ scene, the martingales of the detector and the reasoner for the brightness feature increase above the threshold after $t=2$ seconds. Also, the martingale of the reasoner for precipitation shows a slight variation but does not increase above the threshold. So, we conclude the brightness feature is the cause of the \ac{ood}. In the $HPB$ scene, the precipitation and brightness feature increase to a value out of the training distribution at $t=2$ seconds. The martingale of the detector and both the feature reasoners increase above the threshold. So, we attribute both the features to be the cause of the \ac{ood}. The peaks in the steering plots at $t=2$ seconds are when the DAVE-II \ac{dnn} steering predictions get erroneous, and the decision manager arbitrates the control to the autopilot controller.

The $NR$ scene is of interest for this work, as the scene has a new road segment with different background artifacts (e.g., buildings, traffic lights) that were not in the training distribution. But the precipitation and brightness feature values were within the training distribution. When tested on this scene, the detector martingale instantly increases above the threshold, identifying the scene to be \ac{ood}. However, the martingales of the reasoners for precipitation and brightness only show slight variations without increasing beyond the threshold. The result implies that the scene is \ac{ood}, but a feature other than brightness and precipitation has varied and is responsible for the \ac{ood}.

Further, \cref{fig:change-point} shows the plots of the capability of the  \ac{bvae} detector in identifying changes in the current test image as compared to the previous images in the time series (\emph{problem2}). For these evaluations, we used the $Nom$, $HB$, and extended $HB$ test scenes, which had a length of $50$ seconds each. The $Nom$ and $HB$ scenes are same as the previous setup, but in the extended $HB$ scene, the brightness feature value abruptly increases only for a brief period between $9.5$ seconds and $29.5$ seconds. For these test scenes, we performed the moving average calculation on a sliding window of $M=20$ with the \ac{cusum} parameters of $\omega_{cp}=0.75$ and $\tau_{cp}=1.2$. With these parameters, our detector could identify change points with a short latency of $11$ frames, which translates to one second of inference time for the \ac{av} system.}

\input{results/3stagger}

\subsubsection{Evaluating the Design Approach}
\label{sec:design-evaluation}
\revision{\cref{Table:design-space} illustrates how the proposed heuristics in each step of our approach (\cref{sec:approach-overview}) results in achieving the best detector properties (text highlighted in green). The properties of interest are robustness, minimum sensitivity, and resource efficiency, which are defined in \cref{sec:problem}. We evaluate robustness on the in-distribution scene $Nom$, and two \ac{ood} scenes $HPB$ and $NR$. The $HP$ and $HB$ scenes had variations in either the precipitation value or the brightness value, but they did not have simultaneous variations in both. So, it was suitable to use them to measure the detector's minimum sensitivity towards these features. However, in the $HPB$ scene, both these features varied simultaneously, so it was not suitable for measuring the minimum sensitivity. The resource efficiency was measured across all the scenes. For these evaluations, we have compared the proposed heuristics to an alternate technique in each step. The comparisons are as follows (proposed techniques are underlined): (1) \underline{MIG} vs. ELBO loss function (discussed in \cref{sec:background}), (2) \underline{\acl{bo}} vs. Grid and Random Search, and (3) \underline{Selective} vs. All latent variables for detection. Our evaluations are as follows.

\begin{table}[!t]
\resizebox{\columnwidth}{!}{%
\footnotesize{}
\begin{tabular}{|c|c|c|c|c|c|c|}
\hline
\multicolumn{4}{|c|}{\textbf{Design-time steps of our approach}}                                                                                                                                                                                                                                          & \multicolumn{3}{c|}{\textbf{Detector Properties}}                                                                                                                                                                                                                                         \\ \hline
                                             & \multicolumn{2}{c|}{\textbf{Latent Space Encoding}}                                                                                                &                                                                                                &                                                                                               &                                                                                                &                                                                                          \\ \cline{2-3}
\multirow{-2}{*}{\textbf{Partitioning}}      & \textbf{\begin{tabular}[c]{@{}c@{}}Objective \\ Function\end{tabular}} & \textbf{\begin{tabular}[c]{@{}c@{}}Optimization\\ Algorithm\end{tabular}} & \multirow{-2}{*}{\textbf{\begin{tabular}[c]{@{}c@{}}Selected $\mathcal{L}_d$\\ for Detection\end{tabular}}} & \multirow{-2}{*}{\textbf{\begin{tabular}[c]{@{}c@{}}Robustness\\ F1-score (\%)\end{tabular}}} & \multirow{-2}{*}{\textbf{\begin{tabular}[c]{@{}c@{}}Minimum \\ Sensitivity (\%)\end{tabular}}} & \multirow{-2}{*}{\textbf{\begin{tabular}[c]{@{}c@{}}Execution\\ Time (ms)\end{tabular}}} \\ \hline
{\color[HTML]{009901} }                      & {\color[HTML]{009901} }                                                & {\color[HTML]{009901} \textbf{BO (30,1.4)}}                                        & {\color[HTML]{009901} {[}\textbf{0,2,20,25,29}{]}}                                                     & {\color[HTML]{009901} \textbf{96.98}}                                                                  & {\color[HTML]{009901} \textbf{96}}                                                                      & {\color[HTML]{009901} \textbf{74.09}}                                                             \\ \cline{3-7} 
{\color[HTML]{009901} }                      & {\color[HTML]{009901} }                                                & {\color[HTML]{009901} \textbf{Grid (30,1.4)}}                                      & {\color[HTML]{009901} {[}\textbf{0,2,20,25,29}{]}}                                                     & {\color[HTML]{009901} \textbf{96.98}}                                                                  & {\color[HTML]{009901} \textbf{96}}                                                                      & {\color[HTML]{009901} \textbf{74.09}}                                                             \\ \cline{3-7} 
{\color[HTML]{009901} }                      & \multirow{-3}{*}{{\color[HTML]{009901} \textbf{MIG}}}                           & Random (40,1.5)                                                           & {[}0,17,6,8,7{]}                                                                               & 80.75                                                                                         & 35                                                                                             & 79.15                                                                                    \\ \cline{2-7} 
{\color[HTML]{009901} }                      &                                                                        & BO (30,1.0)                                                               & {[}0,1,6,21,23{]}                                                                              & 95.83                                                                                         & 63                                                                                             & 75.39                                                                                    \\ \cline{3-7} 
{\color[HTML]{009901} }                      &                                                                        & Grid (40,1.0)                                                             & {[}13,28,26,23,0{]}                                                                            & 84.65                                                                                         & 54                                                                                             & 78.85                                                                                    \\ \cline{3-7} 
\multirow{-6}{*}{{\color[HTML]{009901} \textbf{Yes}}} & \multirow{-3}{*}{ELBO}                                                 & Random (30,1.2)                                                           & {[}10,14,21,22,26{]}                                                                           & 94.9                                                                                          & 71                                                                                             & 74.98                                                                                    \\ \hline
                                             &                                                                        & BO (30,1.0)                                                               & All 30                                                                                         & 85.89                                                                                         & 73                                                                                             & {\color[HTML]{FE0000} \textbf{379.47}}                                                            \\ \cline{3-7} 
                                             &                                                                        & Grid (40,1.0)                                                             & All 40                                                                                         & 68.96                                                                                         & 42                                                                                             & {\color[HTML]{FE0000} \textbf{488.85}}                                                            \\ \cline{3-7} 
\multirow{-3}{*}{No}                         & \multirow{-3}{*}{ELBO}                                                 & Random (30,1.2)                                                           & All 30                                                                                         & 89.73                                                                                         & 53                                                                                             & {\color[HTML]{FE0000} \textbf{396.89}}                                                            \\ \hline
\end{tabular}
}
\caption{\revision{\textbf{Design Approach Evaluation}: Evaluations of how the heuristics of our design approach influence the detector properties discussed in \cref{sec:problem}. We evaluated Robustness on the $Nom$, $HPB$, and $NR$ scenes. Minimum sensitivity was evaluated on the $HP$ and $HB$ scenes. Resource efficiency was measured across all the test scenes. The numbers in the optimization algorithm column indicate the selected hyperparameters. Text in green highlights the best detector properties achieved by our approach. Text in red highlights a high detection time.}}
\label{Table:design-space}
\normalsize{}
\vspace{-0.2in}
\end{table}


If the dataset can be partitioned, either \ac{mig} or \ac{elbo} can be used as the objective function with the \ac{bo}, grid, or random search algorithms. Since the dataset can be partitioned, the latent variable heuristic could be applied to select a subset of latent variables, which can be used for detection. As illustrated in \cref{Table:design-space}, the optimization algorithm and objective function combinations resulted in $5$ different \ac{bvae} networks and $5$ latent variables for $\mathcal{L}_d$. Among these, the \ac{bo} and grid algorithms using \ac{mig} resulted in the best detector that had robustness of $96.98\%$, minimum sensitivity of $96\%$, and a detection time of $74.09$ milliseconds. In comparison, the other detectors had low robustness and minimum sensitivity, and a similar detection time.

However, if the dataset cannot be partitioned, then \ac{elbo} is the only objective function that can be used with the \ac{bo}, grid, or random search algorithms. However, without partitioning, the latent variable mapping heuristic cannot be applied. So, all the latent variables for the chosen \ac{bvae} had to be used for detection. Using all the latent variables for detection resulted in a less robust and sensitive detector that took an average of $400$ milliseconds (text in red) as shown in \cref{Table:design-space}. }

\subsection{Detection Results from Competing Baselines}
\label{sec:comparison}
We compare the performance of the \ac{bvae} detector to other state-of-the-art approaches using our \ac{av} example in CARLA. The approaches that we compare against are: (1) Deep-SVDD one-class classifier; (2) \ac{vae} based reconstruction classifier; (3) chain of one-class Deep-SVDD classifiers; and (4) chain of \ac{vae} based reconstruction classifiers. The \ac{vae} network architecture is the same as that of \ac{bvae} (described in \cref{sec:evaluation}) but uses the hyperparameters of $\beta=1$ and $n=1024$. The one-class Deep-SVDD network has four convolutional layers of 32/64/128/256 with $5 x 5$ filters with LeakyReLu activation functions and $2 x 2$ max-pooling, and one fully connected layer with 1568 units. These networks are also trained using a two-learning scheduler, with $100$ epochs at a learning rate $\eta$ = $1$ x $10$\textsuperscript{$-4$}, and $50$ epochs at a learning rate $\eta$ = $1$ x $10$\textsuperscript{$-5$}. Further, we combined two of these classifiers to form a chain of Deep-SVDD classifiers and a chain of \ac{vae} classifiers. In each chain, one classifier is trained to classify images with variations in the values of the brightness feature, and the other is trained to classify images with variations in the values of the precipitation feature. These classifiers are combined using an OR operator. 

For these evaluations, we set resource limits on the python software component (See \cref{fig:carla-model}) for evaluating the processing time and memory usage.  We assigned a soft limit of one CPU core and a hard limit of four CPU cores on each of the components to mimic the settings of an NVIDIA Jetson TX2 board. Further, to measure the memory usage, we used the psutil \cite{rodola2016psutil} cross-platform library.

\subsubsection{Comparing Runtime \ac{ood} Detection}
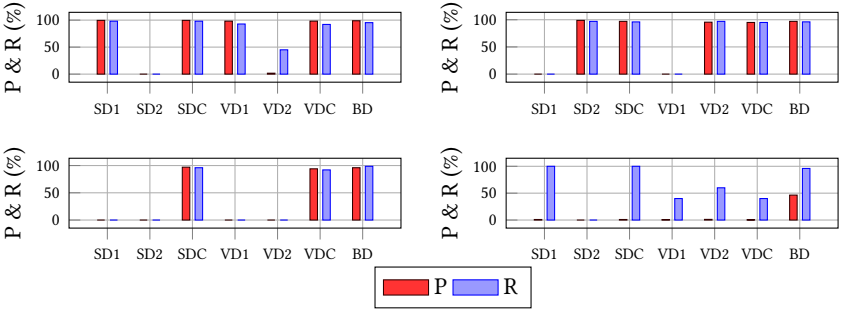
\begin{figure}[!ht]
\begin{tikzpicture}
\begin{groupplot}[group style={group size=2 by 2,horizontal sep = 1.4cm}]
\nextgroupplot[
height=2.5cm, width=6cm,
        bar width=0.1cm,
        ylabel style={yshift=-0.1cm},
    ybar=2pt, area legend,
    x grid style={white!69.0196078431373!black},
xmajorgrids,
y grid style={white!69.0196078431373!black},
ymajorgrids,
    enlargelimits=0.15,
    label style={font=\small},
    tick label style={font=\tiny},
    legend style={at={(0.5,0.1),font=\small},
      anchor=north,legend columns=-1},
    ylabel={P \& R (\%)},
    symbolic x coords={SD1,SD2,SDC,VD1,VD2,VDC,BD},
    xtick=data,
    xtick pos=left,
ytick pos=left,
    nodes near coords align={vertical},
    ]
\addplot[red!20!black,fill=red!80!white] coordinates {(SD1,99.5)(SD2,0)(SDC,99.5)(VD1,98.1)(VD2,1.7)(VDC,98.1)(BD,98.8)};
\addplot[blue!20!blue,fill=blue!40!white] coordinates {(SD1,98)(SD2,0)(SDC,98)(VD1,92.8)(VD2,45)(VDC,92)(BD,95.3)};
\nextgroupplot[
height=2.5cm, width=6cm,
        bar width=0.1cm,
    ybar=2pt, area legend,
    ylabel style={yshift=-0.1cm},
    x grid style={white!69.0196078431373!black},
xmajorgrids,
y grid style={white!69.0196078431373!black},
ymajorgrids,
    enlargelimits=0.15,
    label style={font=\small},
    tick label style={font=\tiny},
    legend style={at={(0.5,-0.55),font=\small},
      anchor=north,legend columns=-1},
    ylabel={P \& R (\%)},
    symbolic x coords={SD1,SD2,SDC,VD1,VD2,VDC,BD},
    xtick=data,
    xtick pos=left,
ytick pos=left,
    nodes near coords align={vertical},
    ]
\addplot[red!20!black,fill=red!80!white] coordinates {(SD1,0)(SD2,98.8)(SDC,97)(VD1,0)(VD2,95.5)(VDC,95)(BD,97)};
\addplot[blue!20!blue,fill=blue!40!white] coordinates {(SD1,0)(SD2,97)(SDC,96)(VD1,0)(VD2,97)(VDC,95)(BD,96.1)};
\nextgroupplot[
height=2.5cm, width=6cm,
        bar width=0.1cm,
        ylabel style={yshift=-0.1cm},
    ybar=2pt,
    x grid style={white!69.0196078431373!black},
xmajorgrids,
y grid style={white!69.0196078431373!black},
ymajorgrids,
    enlargelimits=0.15,
    label style={font=\small},
    tick label style={font=\tiny},
    legend style={at={(0.5,-0.35),font=\small},
      anchor=north,legend columns=-1},
    ylabel={P \& R (\%)},
    symbolic x coords={SD1,SD2,SDC,VD1,VD2,VDC,BD},
    xtick=data,
    xtick pos=left,
ytick pos=left,
    nodes near coords align={vertical},
    ]
\addplot[red!20!black,fill=red!80!white] coordinates {(SD1,0)(SD2,0)(SDC,97)(VD1,0)(VD2,0)(VDC,94)(BD,96)};
\addplot[blue!20!blue,fill=blue!40!white] coordinates {(SD1,0)(SD2,0)(SDC,96)(VD1,0)(VD2,0)(VDC,92)(BD,98.6)};
\nextgroupplot[
height=2.5cm, width=6cm,
        bar width=0.1cm,
    ybar=2pt,area legend,
    ylabel style={yshift=-0.1cm},
    enlargelimits=0.15,
    x grid style={white!69.0196078431373!black},
xmajorgrids,
y grid style={white!69.0196078431373!black},
ymajorgrids,
tick label style={font=\tiny},
label style={font=\small},
    legend style={at={(-0.16,-0.55),font=\small},
      anchor=north,legend columns=-1},
    ylabel={P \& R (\%)},
    symbolic x coords={SD1,SD2,SDC,VD1,VD2,VDC,BD},
    xtick=data,
    xtick pos=left,
ytick pos=left,
    nodes near coords align={vertical},
    ]
\addlegendentry{P}
\addplot[red!20!black,fill=red!80!white] coordinates {(SD1,1)(SD2,0)(SDC,1)(VD1,0.8)(VD2,1.2)(VDC,0.5)(BD,46.4)};
\addlegendentry{R}
\addplot[blue!20!blue,fill=blue!40!white] coordinates {(SD1,100)(SD2,0)(SDC,100)(VD1,40)(VD2,60)(VDC,40)(BD,96)};
\end{groupplot}
\end{tikzpicture}
\setlength{\abovecaptionskip}{2pt}
\caption{\textbf{Precision and Recall}: Evaluations on different scenes: (a) $HP$ - top left, (b) $HB$ - bottom left, (c) $HPB$ - top right and (d) $NR$ - bottom right. The detectors compared are: $SD1$ - Deep-SVDD Precipitation detector, $SD2$ - Deep-SVDD brightness detector, $SDC$ - Deep-SVDD detector chain, $VD1$ - \ac{vae} Precipitation detector, VD2 - \ac{vae} brightness detector, VDC - \ac{vae} detector chain, BD - \ac{bvae} detector. These values were collected by running the detectors on each scene for 20 times.}
\label{fig:PR}
\end{figure}

The false positive and false negative definitions for the $HP$, $HB$, $HPB$, and $NR$ test scenes are defined in \cref{Table:metric-table}. Based on these definitions, the precision and recall of the different detectors for the test scenes are shown in \cref{fig:PR}. In $HP$ and $HB$ scenes, a single feature (precipitation or brightness) value was varied, and the classifier that was not trained on the representative feature of that scene had low true positives. So, the precision and recall of these detectors are mostly zero. In the $HPB$ scene shown in  \cref{fig:PR}-c, both the features were varied, so a single one-class classifier was not sufficient to identify both the feature variations. However, the one-class classifier chains with an OR logic had higher precision in detecting the feature variations in all these scenes. Similarly, the detection and \ac{ood} reasoning capability of the \ac{bvae} detector could identify both the feature variations. 

For the $NR$ scene shown in \cref{fig:PR}-d, which had a new road segment with the precipitation and brightness values within the training distribution, both the one-class classifiers and their chains raise a false alarm. \revision{So, their precision towards detecting variations in these features is low (roughly $1\%$). In contrast, the \ac{bvae} detector alongside the reasoner was able to identify that the \ac{ood} behavior was not because of the precipitation and brightness with a precision of $46\%$.} These results imply that the \ac{bvae} detector can precisely identify if the features of interest are responsible for the \ac{ood}. Whereas a similar reasoning inference cannot be achieved using the other approaches.

\begin{table}[!ht]
\footnotesize
\begin{tabular}{|c|l|l|}
\hline
\textbf{Scene} & \multicolumn{1}{c|}{\textbf{FP}}                                                                                                 & \multicolumn{1}{c|}{\textbf{FN}}                                                                                                        \\ \hline
\textbf{$HP$}    & \begin{tabular}[c]{@{}l@{}}The image is not OOD due to  high precipitation, \\ but it is identified as OOD.\end{tabular}                  & \begin{tabular}[c]{@{}l@{}}The image is OOD due to high rain,  but is \\ identified as in-distribution\end{tabular}                     \\ \hline
\textbf{$HB$}    & \begin{tabular}[c]{@{}l@{}}The image is not OOD due to high \\ brightness, but is identified as OOD.\end{tabular}                & \begin{tabular}[c]{@{}l@{}}The image is OOD due to high brightness, \\ but is identified as in-distribution.\end{tabular}               \\ \hline
\textbf{$HPB$}    & \begin{tabular}[c]{@{}l@{}}The image is not OOD due to high precipitation \\ and high brightness  but, is identified as OOD.\end{tabular} & \begin{tabular}[c]{@{}l@{}}The image is OOD due to high rain  and high\\  brightness, but is identified as in-distribution.\end{tabular} \\ \hline
\textbf{$NR$}    & \begin{tabular}[c]{@{}l@{}}The image is not OOD due to change in road \\ segment, but is identified as OOD.\end{tabular}      & \begin{tabular}[c]{@{}l@{}}The image is OOD due to change in road \\ segment, but is identified as in-distribution.\end{tabular}    \\ \hline
\end{tabular}
\caption{\textbf{Metrics}: Definitions of false positives and false negatives for the $HP$, $HB$, $HPB$, and $NR$ test scenes.}
\normalsize{}
\label{Table:metric-table}
\vspace{-0.2in}
\end{table}

\vspace{-0.12in}

\subsubsection{Comparing Execution Time and Latency}
To emulate a resource constrained setting, we performed the evaluation for execution time and latency on one CPU core. 

\textbf{Execution Time}: As discussed earlier, we selected $5$ latent variables for detection and one latent variable each for reasoning about the precipitation and brightness features. \revision{Two components that mainly contribute to the execution time of the \ac{bvae} detector are: (1) the time taken by the \ac{bvae}'s encoder to generate the latent variables, and (2) the time taken by ICP and martingale for runtime detection. The average execution time using all $30$ latent variables of the \ac{bvae} was $400$ milliseconds, and this was drastically reduced to $74.09$ milliseconds when the $5$ selected latent variables were used. Also, the reasoner only took about $9$ milliseconds as it worked in parallel to the detector.}

\begin{figure}[!h]
\centering
\begin{tikzpicture}
\footnotesize

\begin{groupplot}[group style = {group size = 2 by 1, horizontal sep = 55pt}, width = 6.0cm, height = 5.0cm]
        \nextgroupplot[ 
            height=3cm, width=6.5cm,
        bar width=0.1cm,
    ybar,
    enlargelimits=0.15,
    x grid style={white!69.0196078431373!black},
xmajorgrids,
y grid style={white!69.0196078431373!black},
ymajorgrids,
    legend style={at={(0.5,-0.4),font=\tiny},
      anchor=north,legend columns=-1},
    ylabel style={align=center}, ylabel= Execution\\ Time (ms),
    symbolic x coords={S2,S3,S4,S5},
    xtick=data,
    xtick pos=left,
ytick pos=left,
    nodes near coords align={vertical},
    ]
\addplot [red!20!black,fill=red!80!white]coordinates {(S2,42.6)(S3,42.9)(S4,41.2)(S5,42)};
\addplot coordinates {(S2,42.4)(S3,43.1)(S4,43.4)(S5,42.56)};
\addplot coordinates {(S2,43.3)(S3,44.6)(S4,44.7)(S5,43.3)};
\addplot coordinates {(S2,56.5)(S3,56.1)(S4,57.8)(S5,56.8)};
\addplot coordinates {(S2,56.2)(S3,56.9)(S4,55.3)(S5,55.9)};
\addplot coordinates {(S2,57.2)(S3,57.5)(S4,57.3)(S5,57.6)};
\addplot coordinates {(S2,73.6)(S3,71.4)(S4,72.7)(S5,74.3)};
        \nextgroupplot[height=3cm, width=6.5cm,
        bar width=0.1cm,
    ybar, area legend,
    enlargelimits=0.15,
    x grid style={white!69.0196078431373!black},
xmajorgrids,
y grid style={white!69.0196078431373!black},
ymajorgrids,
    legend style={at={(-0.2,-0.4),font=\tiny},
      anchor=north,legend columns=-1},
   ylabel style={align=center}, ylabel= Avg.\\ Latency \\ (Frames),
    symbolic x coords={S2,S3,S4,S5},
    xtick=data,
    xtick pos=left,
ytick pos=left,
    nodes near coords align={vertical},
    ]
            \addplot[red!20!black,fill=red!80!white] coordinates {(S2,11)(S3,11)(S4,11)(S5,3)};
\addplot coordinates {(S2,9)(S3,10)(S4,9)(S5,4)};
\addplot coordinates {(S2,9)(S3,11)(S4,11)(S5,3)};
\addplot coordinates {(S2,14)(S3,11)(S4,12)(S5,5)};
\addplot coordinates {(S2,13)(S3,15)(S4,13)(S5,6)};
\addplot coordinates {(S2,13)(S3,11)(S4,12)(S5,5)};
\addplot coordinates {(S2,13)(S3,11)(S4,11)(S5,8)};

   \legend{SD1,SD2,SDC,VD1,VD2,VDC,B-VAE}             
    \end{groupplot}

\end{tikzpicture}
\caption{\textbf{Timing Analysis}:(Left) Execution Time in milliseconds and (Right) Detection latency in number of frames, of the different detectors for the $HP$, $HB$, $HPB$ and $NR$ scenes. The detectors are $SD1$ - Deep-SVDD Precipitation detector, $SD2$ - Deep-SVDD brightness detector, $SDC$ - Deep-SVDD detector chain, $VD1$ - \ac{vae} Precipitation detector, $VD2$ - \ac{vae} brightness detector, $VDC$ - \ac{vae} detector chain, $BD$ - \ac{bvae} detector. These values were collected by running the detectors on each scene for 20 times.}
\label{fig:memory}
\vspace{-0.1in}
\end{figure}
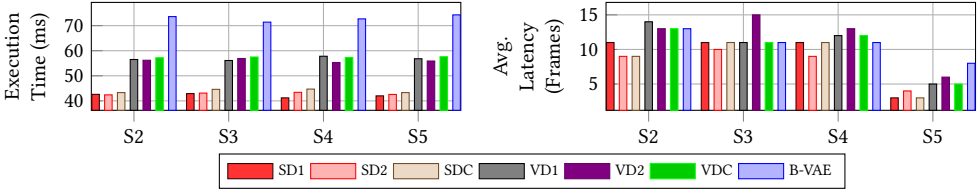

In comparison, the Deep-SVDD classifiers took an average of $41$ milliseconds for detection, and its chain took an average of $43.36$ milliseconds, as shown in \cref{fig:memory}. Also, each of the \ac{vae} based reconstruction classifiers took an average of $53$ milliseconds for detection, and its chain took an average of $57$ milliseconds. The Deep-SVDD and \ac{vae} based reconstruction classifiers performed slightly faster than our detector. 

\textbf{Detection Latency}: \revision{The detection latency in our context is the number of frames between the detection and the occurrence of the \ac{ood}. In our approach, the latency is dependent on the size of the martingale window, which is dependent on the \ac{cps} dynamics and the sampling period of the system as discussed in \cref{sec:icp}. In addition, the selection of the \ac{cusum} threshold also impacts the latency. In these experiments, our \ac{av} traveled at a constant speed of 0.5 m/s, so we used a fixed window size of 20 images. Further, we empirically selected the \ac{cusum} threshold to be $100$. With this configuration, the \ac{bvae} detector had an average latency of  $12.6$ frames, and the reasoners had an average latency of about $11.5$ frames across the $4$ test scenes. In comparison, the Deep-SVDD classifiers and their chain had an average latency of $11$ frames and $11.21$ frames, respectively. Also, each of the \ac{vae} based reconstruction classifiers and their chain had an average latency of $13$ frames and $12.9$ frames, respectively. 

To summarize, all the approaches had similar detection latency, with the Deep-SVDD classifier performing slightly better. The Deep-SVDD classifier and its chain had slightly shorter latency as compared to our detector. However, the \ac{bvae} detector had a lower latency compared to the \ac{vae} based reconstruction classifier and its chain.}

\subsubsection{Comparing Memory Usage}
\label{sec:memory}
Our approach uses a single \ac{bvae} network to perform both detection and reasoning. Specifically, we only utilize the network's encoder instead of both the encoder and the decoder. The average memory utilization of the \ac{bvae} detector was $2.49$ GB, and the reasoners were $0.23$ GB. In comparison, the Deep-SVDD classifiers and their chain utilized an average memory of $3.2$ GB and $6.4$ GB, respectively. Also, the \ac{vae} based reconstruction classifiers and their chain utilized an average memory of $3.6$ GB and $7.2$ GB, respectively. The classifier chains utilized higher memory because two \acp{dnn}s were used for detection. 

To summarize, our approach utilizes lesser memory because: (1) it only requires the encoder of a single \ac{bvae} network for both detection and reasoning; and (2) it utilizes fewer latent variables for detection because it relies on the disentanglement concept. For the \ac{av} example, our detector only required $5$ latent variables as compared to $1024$ latent variables of the \ac{vae} reconstruction classifier and $1568$ activation functions in the embedding layer of the Deep-SVDD classifier.  


\begin{figure*}[!ht]
\setlength{\abovecaptionskip}{0pt}
    \centering
    \includegraphics[width=0.55\textwidth]{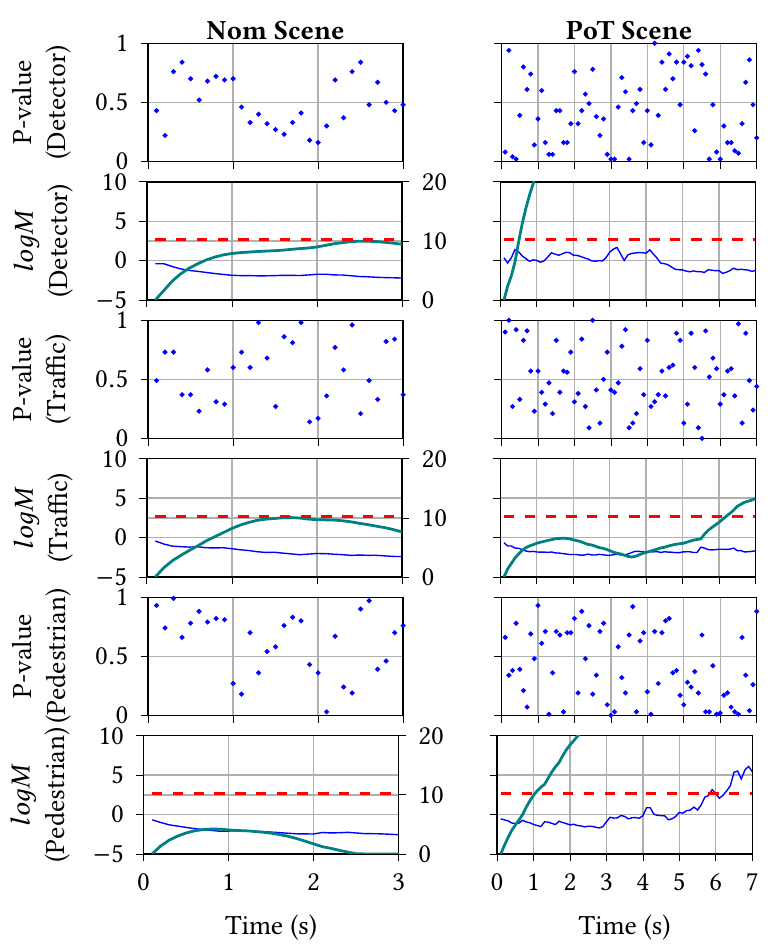}
    \caption{\revision{\textbf{Runtime \ac{ood} Detection on nuImages dataset}: Performance of the \ac{bvae} detector for the $Nom$ (nominal) and $PoT$ (pedestrian or traffic) scenes of the nuImages experiments. For the martingale plots, the solid blue line represents the log of martingale, the solid green lines represent the \ac{cusum} values and the dotted red lines represent the threshold ($\tau$) for \ac{cusum} comparison. Further, the left y-axis shows the log of martingale with range $[-5,10]$, and the right y-axis shows the \ac{cusum} value with range $[0,20]$. The cusum threshold represented in the red dotted lines are plotted to the right y-axis.}}
    \label{fig:nuImage-confidence}
\vspace{-0.12in}
\end{figure*}

\subsection{\ac{ood} Detection Results from nuImages dataset}
\label{sec:nuimage}
\revision{As the second example, we apply our detection approach on a small fragment of the nuImages \cite{nuimage} dataset. We report the preliminary results of our evaluation in this section. 

\textbf{Dataset Overview}: The nuImages dataset is derived from the original nuScenes dataset \cite{caesar2019nuscenes}. The nuImages dataset provides image annotation labels, reduced label imbalance, and a larger number of similar scenes (e.g., a scene with the same road segment but different time-of-day, weather, and pedestrian density values), which makes it suitable for our work. The dataset has $93,000$ images collected at $2Hz$, and it has annotations for foreground objects such as vehicles, animals, humans, static obstacles, moving obstacles, etc. The foreground objects further have additional attributes about the weather, activity of a vehicle, traffic, number of pedestrians, pose of the pedestrians, among others. We demonstrate our \ac{ood} detection capability on scenes with different values of traffic and pedestrian density features. For these experiments, we train the detector with different traffic and pedestrian values and then used it to detect images in which either traffic or pedestrian features are absent (\ac{ood} condition).

\textbf{Data Partitioning}: The training set $\mathcal{T}$ had images from $4$ scenes with varying traffic and pedestrian values. We partitioned $\mathcal{T}$ into two partitions. $P1$ had images with medium and high pedestrian density values in the range $[1,5]$ and more than $5$, respectively. $P2$ had images with medium and high traffic values in the range $[1,5]$ and more than $5$, respectively. The test scenes for this evaluation were nominal ($Nom$) and a pedestrian or traffic ($PoT$) scene. The $Nom$ scene was in distribution, and it had $30$ images in time series captured on a sunny day from a road segment in Singapore city with traffic and pedestrians in the training distribution range. The $PoT$ scene had $70$ images in time series captured from a similar road segment, and it either had traffic or pedestrians, but not both together. For most images in this scene, the traffic density took a value in the training range, but the pedestrian value was close to zero. Further, our test scenes were short as it was difficult to find longer image sequences that belonged to a sunny day and similar road segment.

\textbf{Detector Design}: We applied our approach discussed in \cref{sec:approach-overview} and we used the same \ac{bvae} network structure as discussed in \cref{sec:detector-design}. Using this set up, we selected a \ac{bvae} network with hyperparameters $n$=30 and $\beta$=1.1, that resulted in the maximum \ac{mig} of $0.0006$. We then identified the detector latent variable set $\mathcal{L}_d$ to be $\{L_0,L_7,L_9,L_{22}\}$. Further, we selected latent variables $L_9$ and $L_{22}$ as the reasoners for the traffic density and pedestrian density partitions, respectively. 

\textbf{\ac{ood} Detection}: For runtime \ac{ood} detection, we used a window size $M$=30 for martingale computation and $\omega$=2 and $\tau$=10 for reasoners and detector \ac{cusum} calculations. The p-value and martingale plots for detection and reasoning are shown in \cref{fig:nuImage-confidence}. The detector Martingale for $Nom$ test scene remained low, and all in-distribution images were detected as in-distribution. The martingale of the traffic density reasoner remained low for most images except for $2$ images. We hypothesize the reason for this could be that the vehicle and background blended well that confused the reasoner. The martingale of the pedestrian reasoner remained low throughout the scene. For OOD scene $PoT$, the detector identified $92\%$ ($65/70$) of the images correctly as OOD. However, the first $5$ images were incorrectly detected to be in-distribution because of the detection latency of our window-based martingale approach. Also, as the traffic density was mostly zero throughout the scene, the martingale of traffic density reasoner is mostly flat throughout. But, for pedestrian reasoner, all the images were identified as OOD except the first $9$ images. The false negatives are primarily because of the complexity of images in a scene and the detection latency. After these images, the martingale increases, and the \ac{cusum} also increase above its threshold. In summary, the reasoners worked reasonably well for these scenes but had a slow martingale growth because of several background attributes like sun glare, trees, and traffic lights.

\textbf{Challenges}: We discuss the several challenges of applying our approach to a real-world dataset. The first challenge is the existence of time series images. Although the nuImages dataset provides the notion of a scene, they contain time gaps, especially after partitioning them into train and test the dataset. Second, the labels provided for the dataset images are coarse grain. For example, in nuImages dataset, the semantic label annotations are only limited to high-level foreground objects like pedestrian and cars, and extracting these labels requires significant pre-processing. Another challenge is the complexity of images in real-world datasets. The presence of excess background information such as trees, traffic signals, shadow, reflections, among others, makes the real-world images complex, and it also impacts the information in the latent space. The other challenge is the absence of scenes in which the feature(s) gradually change their values. This makes it difficult to apply our latent variable mapping. Further, finding similar scenes with variations in the specific feature(s) of interest is difficult. For our experiments, we had to perform significant pre-processing to extract short image sequences to perform \ac{ood} detection and reasoning.}

\subsection{Discussion}
\label{sec:discussion}

\begin{figure}[th!]
\begin{tikzpicture}
\centering
\footnotesize
\begin{axis}[
	xlabel= Number of Labels,
	ylabel style={align=center}, ylabel= Memory\\ Usage (GB),
	x grid style={white!69.0196078431373!black},
xmajorgrids,
y grid style={white!69.0196078431373!black},
ymajorgrids,
xtick pos=left,
ytick pos=left,
	width=8cm,height=3cm,
    legend style={at={(1.05,.5)},anchor=west,font=\tiny,cells={align=center}}
    ]

\addplot[color=red,mark=x] coordinates {
	(1, 3.6)
	(2, 7.2)
	(5, 18)
	(10, 36)
	(20, 72)
	(30, 108)
	(38,136.8)
};

\addplot[color=blue,mark=*] coordinates {
	(1, 2.49)
	(2, 2.95)
	(5, 3.64)
	(10, 4.79)
	(20, 7.09)
	(30, 9.12)
	(38,10.96)
};

\legend{VAE-Reconstruction \\ detector chain, $\beta$-VAE detector}
\end{axis}
\end{tikzpicture}
\setlength{\abovecaptionskip}{0pt}
\caption{\textbf{Memory Overhead}: Predicted memory usage of a \ac{vae} based reconstruction chain and a single $\beta$-VAE detector for the nuScenes images. If a \ac{vae} reconstruction detector is used for each of the $38$ nuScenes labels, then the memory usage would linearly grow to $136.8$ GB. However, a \ac{bvae} detector based on our approach would only require $10.96$ GB. These values were computed based on results in \cref{sec:memory}.}
\label{fig:mem-growth}
\vspace{-0.05in}
\end{figure}
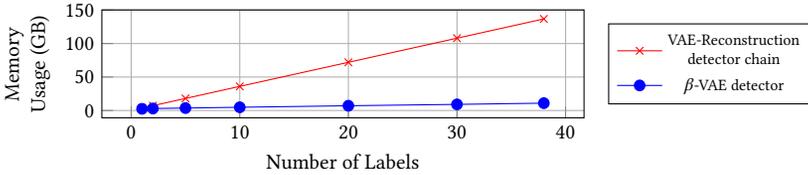

\revision{With \acp{dnn} being widely used in perception pipelines of automotive \ac{cps}, there has been an increased need for \ac{ood} detectors that can identify if the operational test image to the \ac{dnn} is in conformance to the training set. Addressing this problem is challenging because these images have multiple feature labels, and a change in the value of one or more features can cause the image to be \ac{ood}. This problem is commonly solved using a multi-chained one-class classifier with each classifier trained on one feature label. However, as shown by our evaluation in \cref{sec:comparison}, the chain gets computationally expensive with an increased number of image features. So, we have proposed a single \ac{bvae} detector that is sensitive to variations in multiple features and is computationally inexpensive in comparison to the classifier chains. For example, to perform detection on a real-world automotive dataset like nuScenes with $38$ semantic labels, a multi-chained \ac{vae} based reconstruction classifier (discussed in our experiments) would require training $38$ different \ac{vae} \ac{dnn}s as compared to a single \ac{bvae} detector that is presented in this work. A memory projection based on the results in \cref{sec:memory} is shown in \cref{fig:mem-growth}. It shows that the multi-chain classifier will need a memory of $136.8$ GB. In comparison, our approach requires a single network with one or a few latent variables for detection on each label, and this requires only $10.96$ GB of memory.}

\revision{Another related problem motivated in this work is 
the \ac{ood} reasoning capability to identify the most responsible feature(s). The reasoning capability is desirable for the system to decide on the mitigation action it must perform. For example, if the detector can identify a high traffic feature to be the cause of the \ac{ood}, then the system can switch to an alternate controller with lower autonomy. This reasoning capability cannot be achieved using the state-of-the-art chain of one-class classifiers. But the \ac{bvae} detector designed and trained using our approach is sensitive to variations in multiple features. We have evaluated this capability of the \ac{bvae} detector by applying it to several \ac{ood} scenes in CARLA simulation (see \cref{fig:confidence}). Especially the $NR$ scene that had a new road segment and background artifacts that are not in the training set. In this scene, the multi-chained network identified the scene to be \ac{ood} because of the precipitation and brightness features. In comparison, our detector's \ac{ood} reasoning capability was able to identify with a precision of $46\%$ that the cause was not precipitation or brightness.

In addition to the reasoning capability, a detector for a multi-labeled dataset should have minimum sensitivity (defined in \cref{sec:problem}) towards all the feature labels. High minimum sensitivity is needed to detect variations in all the feature labels of the training set images. In our approach, we hypothesize that using the most informative latent variables can provide good minimum sensitivity for the detector towards each feature label. We back this by our results in \cref{Table:design-space}, which illustrates that a detector that used the $5$ most informative latent variables had a high minimum sensitivity to variations in both the precipitation and brightness features (highlighted in green text in the Table). In comparison, the detectors that used all the latent variables for detection had lower minimum sensitivity, as illustrated in the Table.

Besides identifying if the input images are \ac{ood} to the training data, it is necessary to check if they have changed compared to the previous sequence of images in the time series. Such abrupt changes in the current input will increase the system's risk of consequence (e.g., collision) \cite{hartsell2021resonate}, so it is critical to detect them. This problem is necessary to be addressed but has not been given much importance in the \ac{ood} detection literature. In this work, we use the detector's latent variables in a moving window based on \ac{cusum} for detecting abrupt changes in the features of the operational test images. We validated our approach across $3$ different scenes (see \cref{fig:change-point}) for which our detector could accurately identify feature variations with a short latency of about $11$ frames.

Finally, in designing \ac{ood} detectors for \acp{cps}, one needs to take into consideration of the system's dynamics. But this is often not given importance in the existing detection approaches. Only recently, Cai et al. \cite{cai2020real} have proposed an \ac{ood} detection mechanism on time series images that consider the system dynamics. As discussed in \cref{sec:anomaly-detection}, we rely on their \ac{icp} and martingale framework approach for runtime detection over a short sliding window of images. The sliding window size needs to be adjusted based on the operational system dynamics like runtime sampling rate and the speed at which the system is traveling. }

\section{Related Work}
\label{sec:RW}

\revision{There has been significant ongoing research to handle the brittleness and susceptibility of \ac{dnn}s. We have grouped the existing approaches into different classes and briefly discuss them below.}

\revision{\textbf{Domain Adaptation} involves transferring knowledge between a labeled source domain and an unlabeled target domain ~\cite{pan2009survey}. The key idea is to learn domain invariant features of the training data by providing less importance to dataset biases. This is also referred to as transductive transfer learning and is used when the train and the target tasks remain the same while the domains are different. Altering the \ac{dnn} structure has been one of the approaches that have been used for feature transferability. One such example is the Deep Adaptation Network (DAN)~\cite{long2015learning} that allows for feature transferability in the task specific layers of \acp{dnn} network while reducing domain related information. Biasing the training objective to learn the domain invariant feature(s) has been the other widely adopted approach. For example, Heinze-Deml \emph{et al.}~\cite{heinze2021conditional} proposes a conditional variance penalty-based training loss function to learn domain invariant features. Although domain adaptation has been widely used, an assumption on the prior distributions of target domains is restrictive for practical applications like \acp{cps}~\cite{volpi2018generalizing}. Further, negative transfer of knowledge between the source and target domains has been a common problem~\cite{wang2019characterizing}.}
    
\revision{\textbf{Confidence Estimation} involves estimating the confidence in the inputs to a \ac{dnn}. There have been several approaches to estimating confidence. The first approach involves adding a confidence branch at the logits before the softmax layer ~\cite{devries2018learning}. The second approach involves using Bayesian Networks for representing uncertainties in the \acp{dnn} ~\cite{mackay1992practical,neal2012bayesian}. Ensembles of \acp{dnn} have been another approach used for estimating the confidence in the inputs ~\cite{lakshminarayanan2016simple,geifman2018bias}. All these approaches mostly require making changes to the \ac{dnn} or will require training model ensembles. Recently, ~\citeauthor {jha2019attribution} ~\cite{jha2019attribution} have proposed the attribution based confidence (ABC) metric that does not require access to training data or does not need training model ensembles. It is computed by sampling in the neighborhood of high-dimensional data and then computing a score for its conformance. The metric looks robust, and it does not require access to the training data at runtime, which is an advantage compared to our approach. But the impact of input feature correlations and interactions on the attribution over features require future investigation ~\cite{sundararajan2017axiomatic}.  
}

\revision{\textbf{Identifying Distribution Shifts} involves identifying if the test observation has shifted from the training distribution. Probabilistic classifiers like \ac{gan} and \ac{vae} have been widely adopted for identifying shifts in the test observations ~\cite{ruff2018deep,cai2020real,denouden2018improving,an2015variational}. Our work belongs to this class, and different approaches in this class are discussed in \cref{sec:one-class}. }

\revision{In contrast to these approaches, we formulate the problem as a multi-label \ac{ood} detection problem, and in addition to identifying an \ac{ood} condition, we also find the feature(s) causing it.}

\subsection{Probabilistic One-class Classifiers}
\label{sec:one-class}

\subsubsection{Adversarial Networks}
\acp{gan} have emerged as the leading paradigm in performing un-supervised and semi-supervised tasks like classification and \ac{ood} detection ~\cite{akcay2018ganomaly,zenati2018efficient}. \ac{gan} has been shown to outperform classical \ac{ood} techniques on benchmark datasets  ~\cite{zenati2018efficient}. This network is used along with an adversarial training loss function to perform unsupervised \ac{ood} detection ~\cite{akcay2018ganomaly}. Further, Adversarial Variational Bayes (AVB), a training technique to improve the inference model of \ac{vae} ~\cite{mescheder2017adversarial}. It combines \acp{vae} and \acp{gan} to rephrase the maximum-likelihood problem of the \ac{vae} as a two player game. Although \ac{gan} has performed well in detecting \ac{ood} data, there are several problems in using them ~\cite{creswell2018generative, vu2019anomaly}. These problems are (1) training complexity because of the instability between the generator and discriminator networks, and (2) mode collapse problem that results in the generator only producing similar samples or, in the worst-case only a single sample. 


\revision{Adversarial Autoencoder \cite{makhzani2015adversarial} is a probabilistic network that uses the reconstruction capability of an autoencoder along with the adversarial training procedures of a \ac{gan}. For example, the reconstruction error is combined with the likelihood in the latent space to perform anomaly detection ~\cite{beggel2019robust}. Vu, Ha Son \emph{et al.} ~\cite{vu2019anomaly} have proposed a Dual Adversarial Autoencoder network that uses two autoencoders and discriminators to perform anomaly detection. However, training Adversarial Autoencoders is expensive as it requires training both the autoencoder network and the adversarial network.}

\subsubsection{Variational Autoencoders (VAE)}
\ac{vae} is the other unsupervised probabilistic generative model that has been used because of its capability to learn latent space of the input data ~\cite{kingma2013auto}. The latent variables generated by these models are used to learn the correlation between the features of the input data ~\cite{liu2019latent}. The \ac{vae} based \ac{ood} approaches can be classified into two categories.

\textbf{Reconstruction based techniques} use the normalized difference between each point of the input data ($x$) and the reconstructed data ($x'$) generated by a \ac{vae}. For example, the authors in ~\cite{richter2017safe,cai2020real} have used the pixel-wise mean square distance between the input image and the reconstructed image as the metric for anomaly detection. As an extension, the reconstruction probability that uses the stochastic latent variables of \ac{vae} has been used to compute a probabilistic anomaly score ~\cite{an2015variational}. Although being widely used, this approach can be error-prone when the \ac{ood} samples lie on the boundary of the training distribution ~\cite{denouden2018improving}.

\textbf{Latent space based techniques} use distance and density-based metrics on the latent space generated by the encoder of \ac{vae} to detect \ac{ood} observations. For example, Denouden \emph{et al.} ~\cite{denouden2018improving} compute the Mahalanobis distance between the latent distributions of the test image and the mean vector of the images in the training set, which is also combined with the reconstruction loss to detect \ac{ood} images. The authors in ~\cite{vasilev1806q} evaluate the latent space with different metrics like Euclidean distance and Bhattacharyya distance for \ac{ood} detection. Although being used for \ac{ood} detection ~\cite{denouden2018improving,vasilev1806q}, there is a known problem, that the generated latent variables are unstructured, entangled, and lack the ease of understanding ~\cite{klys2018learning}. To address this problem, there has been significant research in structuring and disentangling the latent space ~\cite{bengio2013representation,higgins2016beta,chen2018isolating,mathieu2019disentangling}, which is discussed in the following section.

\subsection{Disentangling Latent Representations}
\revision{Learning a disentangled latent space have been shown to be beneficial for several tasks like pose invariant face recognition ~\cite{tran2017disentangled,peng2017reconstruction}, video predictions ~\cite{hsieh2018learning}, and anomaly detection ~\cite{wang2020oiad}. While there are several approaches to learning a disentangled latent space ~\cite{bengio2013representation}, of particular interest to this work are approaches that use the \ac{vae} structure. Recently, variants of a \ac{vae} like FactorVAE ~\cite{kim2018disentangling}, \ac{bvae} ~\cite{higgins2016beta}, and $\beta$-TCVAE ~\cite{chen2018isolating} have been used. Among these, \ac{bvae} has been widely used because it provides a single $\beta$ gating hyperparameter that can be used control the latent space disentanglement ~\cite{higgins2016beta}. Mathieu, Emile \emph{et al.} ~\cite{mathieu2019disentangling} illustrates that a \ac{bvae} generates a lower overlap of the latent variables as compared to the original \ac{vae}, and illustrates how different values of $\beta$ impacts the latent variable overlap. Further, Locatello \emph{et al.} ~\cite{locatello2018challenging} have illustrated with experiments that network inductive biases are necessary to achieve unsupervised disentanglement.    

In recent years \ac{bvae} is used for \ac{ood} detection tasks because of its ability to generate a disentangled latent space. In our previous work ~\cite{sundar2020out}, we have used the tuning capability of a \ac{bvae} to make it sensitive to different image features (e.g., traffic density, pedestrians) and then used it to detect changes in those features. Graydon \emph{et al.} ~\cite{graydon2018novelty} have used the latent variables generated by a \ac{bvae} along with Gaussian mixture models to identify \ac{ood} on cancer datasets. A combination of the reconstruction error and the distances among the latent variables of a \ac{bvae} is used as the anomaly score ~\cite{li2020anomaly}. Further, a \ac{bvae} is shown to perform better for anomaly detection on MRI datasets as compared to a classical \ac{vae} ~\cite{zhou2020unsupervised}. However, these approaches either randomly select the $\beta$ values or manually tune them, which can be expensive and time consuming. In contrast, our work provides a \acl{bo} based heuristic to select the $\beta$ value.  

Further, to measure the level of disentanglement, several methods and metrics are available. Visualization of reconstructions by inspecting latent traversals has been a simple method that has been widely used. But visualization alone is not sufficient, as discussed in \cref{sec:intro}. So, several qualitative and quantitative metrics have been proposed for this task. A classifier based supervised learning metrics ~\cite{higgins2016beta,kim2018disentangling,grathwohl2016disentangling} have been designed when all the underlying generative factors of an image are known. For example, the BetaVAE metric \cite{higgins2016beta} builds a linear classifier that predicts the index of a fixed factor with variations. However, this metric has a drawback of axis alignment, and this is overcome by the FactorVAE metric ~\cite{kim2018disentangling} by using a majority vote classifier. Also, these metrics require building and tuning the sensitivity of the classifiers, which is difficult to design and train. This problem is overcome by the \acl{mig} ~\cite{chen2018isolating} metric, which is a quantitative metric based on the mutual information between the features and latent variables. In this work, we have used \ac{mig} to measure the level of latent space disentanglement.}

\section{Conclusions and Future Work}
\label{sec:conclusion}
In this paper, we proposed a design approach to generate a partially disentangled latent space and learn an approximate mapping between the latent variables and features for \ac{ood} detection and reasoning. To generate the disentangled latent space, we use a \ac{bvae} network and tune its hyperparameter ($n$, $\beta$) using a \acl{bo} heuristic. Next, we perform a latent variable mapping to identify the most informative latent variables for detection and identify latent variable(s) which are sensitive to specific features and use them for reasoning the \ac{ood} problem. We evaluated our approach using an \ac{av} example in the CARLA simulation and illustrated the preliminary results from the nuImages dataset. Our evaluation has shown that the detector designed using our approach has good robustness, minimum sensitivity, and low execution time. The detector could also detect \ac{ood} images and identify the most likely feature(s) causing it.

Future extensions and applications of the proposed approach include: (1) improving the latent variable mapping heuristic to perform robust correspondence between the features and latent variables, (2) explore alternate metrics such as Wasserstein distance for latent variable mapping, (3) apply the approach to real-world datasets and research \ac{cps} testbeds like DeepNNCar ~\cite{ramakrishna2019augmenting}, and (4) use the anomaly detection results to perform higher-level decision making such as controller selection or enactment of contingency plans.

\section*{Acknowledgements}
We thank the anonymous reviewers of our journal for their insightful comments and valuable suggestions. This work was supported by the DARPA Assured Autonomy program, the Air Force Research Laboratory, and in part by MoE, Singapore, Tier-2 grant number MOE2019-T2-2-040. Any opinions, findings, and conclusions, or recommendations expressed in this material are those of the author(s) and do not necessarily reflect the views of DARPA, AFRL, or MoE Singapore.
\vspace{-0.1em}

\bibliographystyle{ACM-Reference-Format}
\bibliography{main.bib}

\newpage

\end{document}